\documentclass[11pt]{article}

\usepackage[final]{acl}

\usepackage{times}
\usepackage{latexsym}

\usepackage[T1]{fontenc}

\usepackage[utf8]{inputenc}

\usepackage{microtype}
\usepackage{multirow}
\usepackage{booktabs}
\usepackage{amssymb}
\usepackage{amsmath}
\usepackage{graphicx}
\usepackage{longtable}
\usepackage{array}
\usepackage{tcolorbox}
\usepackage{cuted}  
\usepackage{colortbl}
\usepackage{tabularx}
\usepackage{rotating}
\usepackage{listings}
\usepackage{enumitem}
\lstdefinestyle{prompt}{
    basicstyle=\ttfamily\small,
    rulecolor=\color{gray},
    breaklines=true,
    breakatwhitespace=true，
    breakautoindent=false,
    breakindent=0pt
}
\usepackage[table,xcdraw]{xcolor}

\definecolor{good}{HTML}{228B22} 
\definecolor{bad}{HTML}{8B0000} 
\usepackage{inconsolata}

\usepackage{graphicx}

\title{TheraAgent: Self-Improving Therapeutic Agent for Precise and Comprehensive Treatment Planning}

\author{
 \textbf{Junkai Li\textsuperscript{1,2}},
 \textbf{Yunghwei Lai\textsuperscript{1,2}},
 \textbf{Tianyi Zhu\textsuperscript{3}},
 \textbf{Zheng Long Lee\textsuperscript{1,2}},
\\
 \textbf{Weizhi Ma\textsuperscript{2}\thanks{Correspondence to Weizhi Ma (mawz@tsinghua.edu.cn), Yang Liu (liuyang2011@tsinghua.edu.cn).}},
 \textbf{Yang Liu\textsuperscript{1,2}\footnotemark[1]}
\\
 \textsuperscript{1}Dept. of Comp. Sci. \& Tech., Institute for AI, Tsinghua University, Beijing, China \\
 \textsuperscript{2}Institute for AI Industry Research (AIR), Tsinghua University, Beijing, China \\
 \textsuperscript{3}School of Computing, National University of Singapore, Singapore
}

\usepackage[strings]{underscore}
\begin{document}
\maketitle
\begin{abstract}
Formulating a treatment plan is inherently a complex reasoning and refinement task rather than a simple generation problem. However, existing large language models (LLMs) mainly rely on one-shot output without explicit verification, which may result in rough, incomplete, and potentially unsafe treatment plans. To address these limitations, we propose \textbf{TheraAgent}, an agentic framework that replaces one-shot generation with an iterative \textit{generate-reflect-refine} pipeline. Inspired by the iterative revision patterns commonly observed in expert clinical practice, our framework progressively \textit{reflects} on clinical criteria and transforms coarse drafts into precise, comprehensive and safer therapeutic regimens. To facilitate the critical \textit{reflection} component, we introduce \textbf{TheraJudge}, a treatment-specific evaluation module integrated into the inference loop. Experiments show TheraAgent achieves state-of-the-art results on HealthBench in treatment planning task, leading in Accuracy and Completeness. In expert evaluations, it attains an 86\% win rate against physicians, with superior Targeting and Harm Control. Moreover, the high consistency between TheraJudge and HealthBench evaluation indicates the reliability of our framework.

\end{abstract}

\section{Introduction}
\begin{figure}[t]
\centering
\includegraphics[width=\columnwidth]{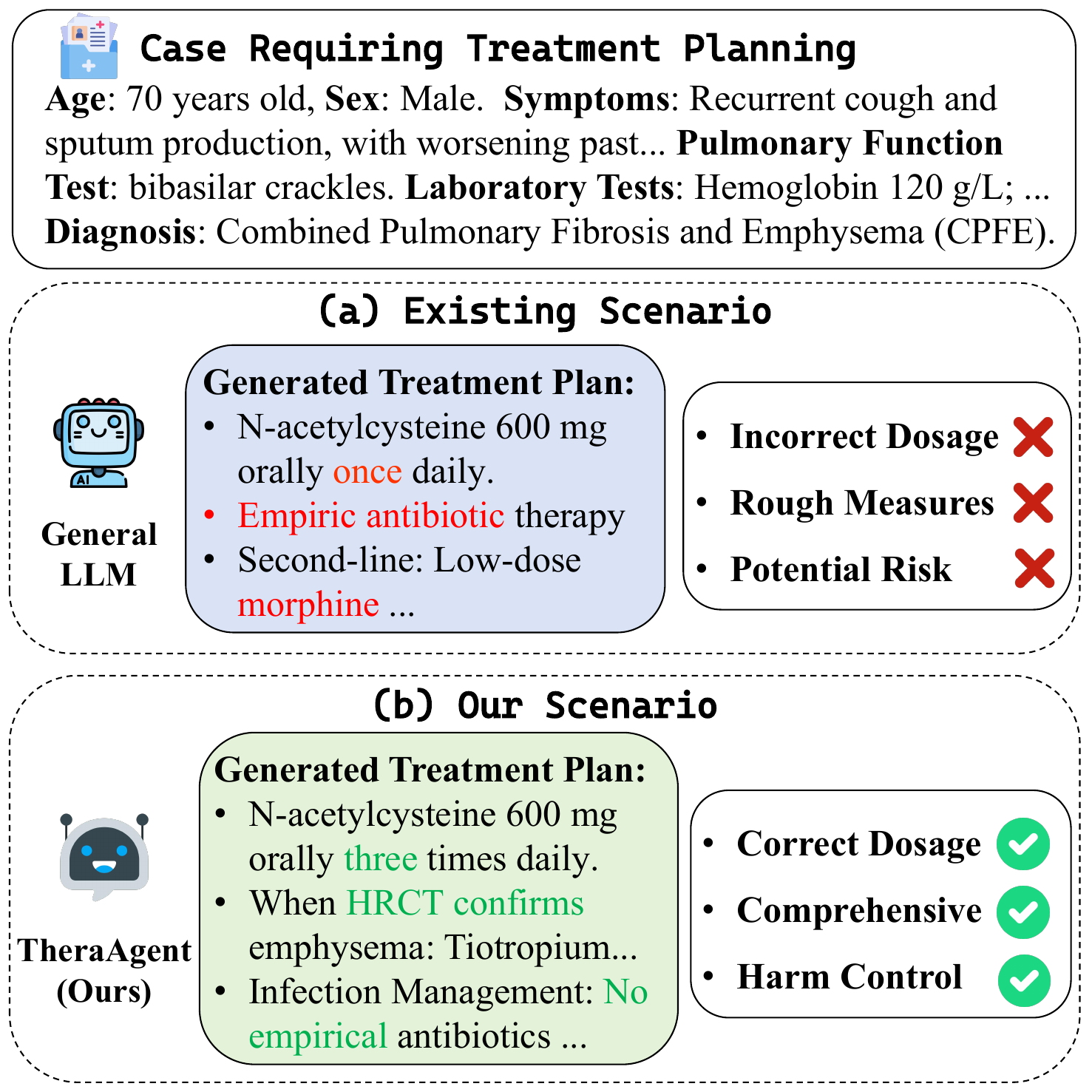}
\vspace{-18pt}
\caption{Comparison of treatment plan generation scenarios.
(a) General-purpose models frequently produce suboptimal treatment plans, characterized by imprecise information, omitted key components, and potential safety risks.
(b) \textbf{TheraAgent} mitigates these limitations by generating more accurate, comprehensive, and clinically reliable plans, with expert-validated annotations.
}
\label{fig:motivation}
\vspace{-12pt}
\end{figure}

Treatment planning is a critical part in real-world medical decision-making, where physicians translate diagnostic conclusions into concrete therapeutic actions to optimize patient outcomes~\cite{effectiveness_of_chat}. It is inherently challenging, as we hypothesize that, rather than relying solely on one-shot generation, effective treatment planning benefits from iterative reasoning and structured refinement, including precise medication selection and dosing, comprehensive therapeutic coverage, and principled harm control~\cite{Evaluation_of_large}. With recent advances in large language models (LLMs), there is growing interest in their potential to support medical tasks, driven by their strong capabilities to incorporate relevant context, perform language comprehension, generate coherent text~\cite{toward_expert_level,a_generalist_medical,toward_accurate_differential,towards_conversational_diag}.

However, directly applying general-purpose or even medically fine-tuned LLMs to treatment planning remains challenging~\cite{performance_analysis_of}. As illustrated in Figure~\ref{fig:motivation}, existing models tend to generate rough, incomplete, and potentially unsafe plans due to their one-shot mechanism without explicit verification. Despite the question, existing research for treatment planning remains limited. Many existing efforts are tailored to specific diseases and thus fail to generalize across clinical departments~\cite{ai_driven_clinical, exploring_the_role, zhang2025large, mohammed2025developingartificialintelligencetool}, while more general approaches often struggle to ensure precision and completeness due to the absence of iterative refinement mechanisms~\cite{hsu2025medplan, yang2025zero}. Furthermore, the evaluation of treatment plans frequently rely on oversimplified metrics (e.g., BLEU) or vanilla LLM-based scores that overlook critical dimensions~\cite{gao2025txagent, hasan2025clinllmsafetyconstrainedhybridframework}.

To bridge these gaps, we propose \textbf{TheraAgent}, an agentic framework that regards treatment planning as an iterative \textit{generate–reflect–refine} process. Inspired by the iterative revision patterns commonly observed in expert clinical practice, TheraAgent enables continuous self-correction and progressive improvement, leading to more precise, complete, and safer treatment plans. To support this process without costly human intervention, we introduce \textbf{TheraJudge}, a treatment-specific evaluation module, which aligns automated feedback with clinical criteria by assessing plans on dimensions including accuracy, targeting, completeness, and safety.

We evaluate TheraAgent on treatment-related cases in HealthBench~\cite{arora2025healthbenchevaluatinglargelanguage}, achieving a state-of-the-art overall score and obtains the highest scores on Accuracy and Completeness, exceeding the second-best model by 2.91 points and 4.43 points. Its effectiveness is further shown in blinded medical expert evaluations on real-world clinical cases, where TheraAgent achieves an 86\% win rate against human physicians, with marked win rate of 69\% in Targeting, 71\% in Completeness, and 51\% in Harm Control. We further conduct an agreement analysis between TheraJudge and HealthBench evaluation, where TheraJudge achieves a \textit{Pearson correlation} of 0.71, substantially outperforming other automatic metrics.

Our contributions are summarized as follows:
\vspace{-0.5em}
\begin{itemize}[left=0cm, itemsep=2pt, parsep=0pt]
    \item We propose \textbf{TheraAgent}, an agentic framework that considers treatment planning as an iterative reasoning process. Its \textit{generate–reflect–refine} pipeline enables effective self-improvement, yielding more precise, comprehensive, and safer treatment plans.
    \item We introduce \textbf{TheraJudge}, a clinically aligned internal critic within TheraAgent that assesses treatment plans along dimensions such as \textit{Accuracy} and \textit{Completeness}. It also shows strong potential as a standalone evaluator for treatment planning, as reflected by its high level of agreement with HealthBench evaluation.
    \item Extensive experiments show that TheraAgent achieves state-of-the-art performance on treatment-related HealthBench cases and attains a dominant 86\% win rate against physicians in blinded expert evaluations on real-world cases.
\end{itemize}
    
\begin{figure*}[h]
\centering
\includegraphics[width=\textwidth]{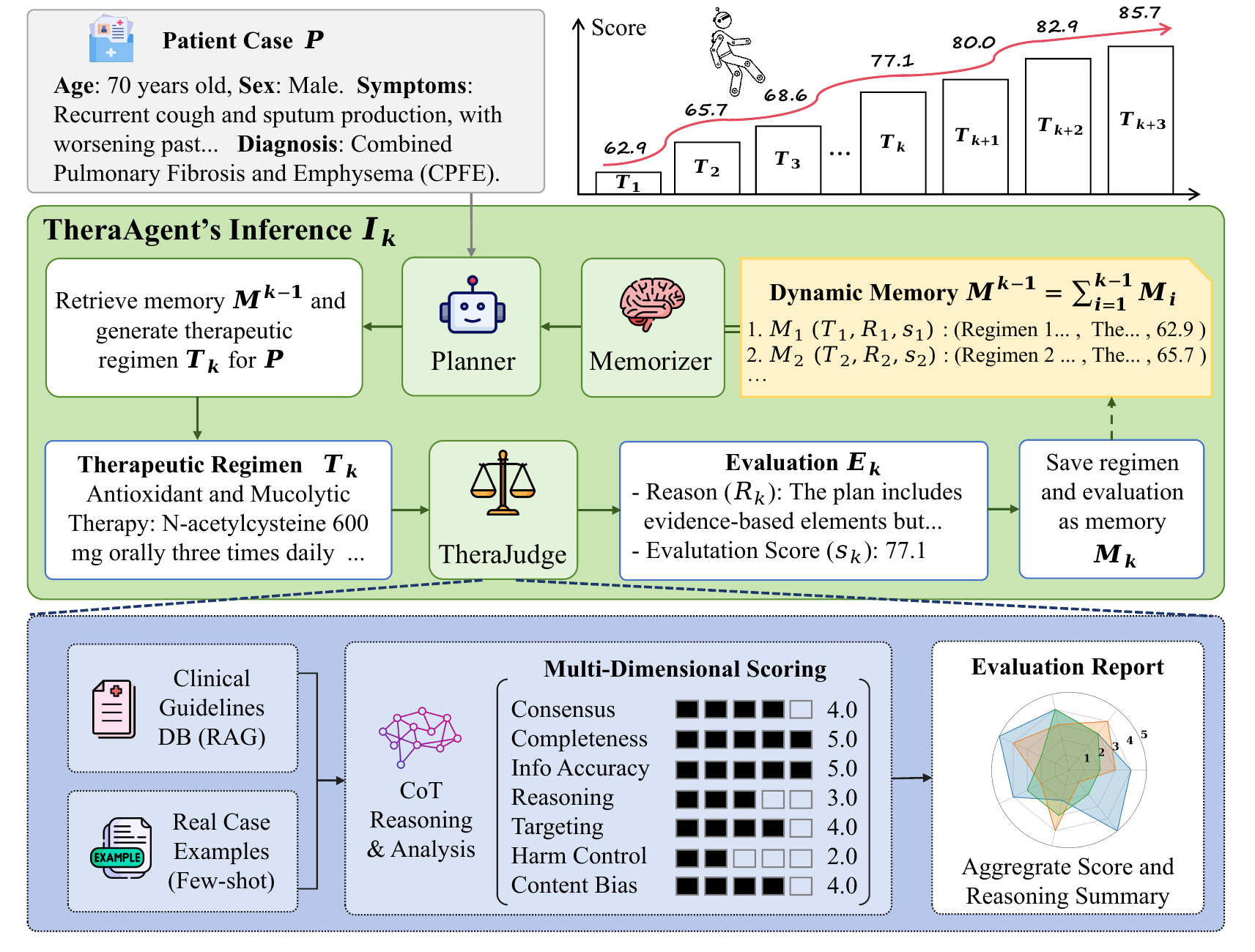}
\vspace{-2em}
\caption{\textbf{Overview of the TheraAgent framework.}
TheraAgent performs treatment planning through a self-improving inference pipeline. Given a patient case $P$, the Planner generates a therapeutic regimen $T_k$ at iteration $k$, which is subsequently assessed by TheraJudge that gives multi-dimensional scores using RAG and Few-shots. The generated schedule and its evaluation are incorporated into the Memorizer to form $M^k$, which enables improved retrieval for the Planner in subsequent iterations and guides refined schedules ($T_{k+1}$) through in-context learning.}
\label{fig:method}
\vspace{-0.5em}
\end{figure*}

\section{Related Works}

\paragraph{Treatment Planning Generation}
Compared to diagnosis-related tasks, research on LLM-based treatment planning remains limited. Existing work includes agent-based or staged generation frameworks aligned with clinical workflows named MedPlan~\cite{hsu2025medplan}, treatment-specific fine-tuning approaches for therapeutic reasoning named TxAgent~\cite{gao2025txagent}, and general medical post-training or test-time optimization methods that improve overall clinical reasoning without explicitly targeting treatment planning named MedCritical~\cite{su2025medcritical} and FineMedLM-o1~\cite{yu2025finemedlmo1}. Prior studies have also examined the evaluation of medical LLMs, typically relying on task-level accuracy or coarse LLM-based scores~\cite{sousa2025performance}.
Despite these efforts, existing approaches often struggle to produce  accurate, complete, and safety-aware treatment regimens, while current evaluation methods remain overly simplified and may fail to capture the multi-dimensional clinical quality required for real-world treatment planning.

\paragraph{Self-improving Agent}
Recent work has demonstrated that agentic and self-improving approaches can improve LLM output quality by enabling iterative refinement. General agent frameworks show that incorporating planning, memory, or reflection leads to more robust and coherent outputs~\cite{huang2024queryagent, liu2025cer, wang2024metareflection, yan2024erm, yin2024godelagent}. In the medical domain, agent-based systems similarly improve reliability and quality through feedback-guided workflows, primarily for diagnosis tasks~\cite{rose2025meddxagent, diao2025guidebench}. Complementary studies on multi-agent evaluation further support iterative refinement as an effective mechanism for assessing and improving reasoning quality \cite{zhao2024autoarena}.
However, these methods are largely designed for general-purpose reasoning or diagnosis tasks, and have not been adapted to the treatment planning setting. Consequently, the application of self-improving agents to generating and refining precise and complete treatment plans remains underexplored. 

\section{Problem Formulation}
Given a patient case
$
\mathbf{P} = (\mathbf{d}, \mathbf{s}, \mathbf{y}),
$
where $\mathbf{d}$ denotes basic clinical information, $\mathbf{s}$ denotes symptoms and clinical findings, and $\mathbf{y}$ denotes the confirmed diagnosis, the goal of treatment planning is to generate a treatment plan
$
\mathbf{T}
$
along with an explicit reasoning trace
$
\mathbf{c},
$
such that the resulting plan is clinically interpretable.
Unlike closed-form prediction tasks, treatment planning operates over an open-ended and combinatorial action space, where valid solutions must simultaneously satisfy multiple clinical requirements. We characterize the quality of a treatment plan as a multi-dimensional function:

\vspace{-0.5em}
\begin{equation}
Q(\mathbf{T} \mid \mathbf{P}) =
\sum_{i=1}^{N} q_i(\mathbf{T} \mid \mathbf{P}),
\end{equation}
where each $q_i$ corresponds to a clinical dimension, such as Accuracy, Targeting, Completeness, and Harm Control. 

This formulation highlights the intrinsic difficulty of treatment planning: high-quality solutions require the model to balance precise medical decisions, comprehensive therapeutic coverage, and safety-aware reasoning under an open-ended space. Consequently, effectively addressing this problem calls for iterative reasoning and refinement mechanisms that can progressively identify and correct deficiencies across multiple clinical dimensions.

\vspace{-0.5em}
\section{TheraAgent}

\subsection{Overview}

To address the above challenges, we propose \textbf{Self-improving therapeutic agent (TheraAgent)}, an agentic workflow that performs iterative optimization over treatment plans via structured feedback. As illustrated in Figure~\ref{fig:method}, TheraAgent consists of three interacting components: \emph{Planner}, \emph{Memorizer}, and \emph{TheraJudge}.

Given a patient case, the Planner generates a candidate treatment plan conditioned on both the current input and historical feedback stored in the Memorizer. The generated schedule is then evaluated by TheraJudge, which provides clinically grounded, multi-dimensional feedback, including detailed rationales and an overall quality score. These evaluation results are stored in the Memorizer and used to guide subsequent generations.


\subsection{Planner}

The \textbf{Planner} serves as the generative module of TheraAgent, responsible for producing treatment plans conditioned on the patient case and historical feedback. At iteration $k$, given a patient case $\mathbf{P}$ and the accumulated memory state $\mathcal{M}^{(k-1)}$, the Planner generates the $k$-th candidate treatment plan $\mathbf{T}_k$ together with its corresponding intermediate reasoning process $\mathbf{c}_k$, formalized as:
\begin{equation}
(\mathbf{T}_k, \mathbf{c}_k)
= f_{\theta}\big(\mathbf{P}, \mathcal{M}^{(k-1)}\big),
\label{eq:planner}
\end{equation}
where $f_{\theta}$ denotes the underlying inference model parameterized by $\theta$.

Unlike one-shot generation, the Planner follows an iterative refinement paradigm by explicitly conditioning on previously generated schedules and their evaluation rationales and scores stored in the Memorizer. This feedback guides the model to identify and correct deficiencies from earlier iterations, such as incomplete coverage, suboptimal clinical targeting, or potential safety risks, thereby transforming treatment plan generation into a progressive optimization process. 

\subsection{TheraJudge}
Each candidate treatment plan generated by the Planner is subsequently evaluated by \textbf{TheraJudge}, which serves as a clinical judging model providing structured, multi-dimensional feedback. Given a patient case $\mathbf{P}$ and the $k$-th candidate schedule $\mathbf{T}_k$, TheraJudge produces evaluation rationales $\mathbf{R}_k$, dimension-wise scores $\{q_{k,i}\}_{i=1}^{N}$ along clinical axes, and an aggregated score $\mathbf{s}_k$, formalized as:

\vspace{-0.5em}
\begin{equation}
(\mathbf{R}_{k}, \{q_{k,i}\}_{i=1}^{N},\mathbf{s}_{k}) 
= g_{\phi}\big(\mathbf{P}, \mathbf{T}_{k}\big),
\label{eq:therajudge}
\end{equation}
where $g_{\phi}$ denotes the TheraJudge model. The final score $\mathbf{s}_k$ is computed as a weighted sum of the individual dimensions, with weights reflecting practical clinical priorities such as Consensus and Completeness. The evaluated tuple $(\mathbf{T}_k, \mathbf{R}_k, \mathbf{s}_k)$ is then stored in the Memorizer to update the memory state $\mathcal{M}^{(k)}$, providing explicit and structured feedback that guides subsequent iterations of treatment plan refinement.

There are three parts in TheraJudge.
\textbf{Knowledge-grounded evaluation via RAG.}
TheraJudge employs a RAG module to retrieve authoritative guidelines based on the patient case and proposed schedule. The retrieval corpus consists of a large-scale collection of real-world clinical guidelines and medical literature, including over 600 documents curated from an authentic website\footnote{https://seleguide.yiigle.com/webs/Knowledge}. By grounding assessments in external medical knowledge, the framework reduces hallucinations and enhances adherence to safety standards.

\textbf{Stabilization via few-shot clinical exemplars.}
To ensure consistent scoring, TheraJudge utilizes few-shot prompting with expert-annotated clinical cases. These exemplars calibrate the model’s judgment, reducing scoring variance compared to zero-shot approaches. This stability is essential for providing the reliable feedback necessary for effective iterative optimization.

\textbf{Multi-dimensional clinical scoring.}
Instead of a single holistic score, TheraJudge performs multi-dimensional assessments covering completeness, safety, and consensus adherence. This mirrors real-world clinical decision-making and provides fine-grained feedback. The resulting reasoning traces are stored in the Memorizer, offering explicit signals that guide the Planner to address specific weaknesses in subsequent iterations.

\begin{table*}[h]
\centering
\resizebox{0.9\textwidth}{!}{%
\begin{tabular}{@{}lcccccccc@{}}
\toprule
 & \multicolumn{1}{c|}{} & \multicolumn{4}{c|}{\textbf{Theme}} & \multicolumn{3}{c}{\textbf{Axis}} \\ \cmidrule(l){3-9} 
\multirow{-2}{*}{\textbf{Model}} & \multicolumn{1}{c|}{\multirow{-2}{*}{\textbf{Overall}}} & \textbf{\begin{tabular}[c]{@{}c@{}}Global\\ Health\end{tabular}} & \textbf{Hedging} & \textbf{\begin{tabular}[c]{@{}c@{}}Context\\ Seeking\end{tabular}} & \multicolumn{1}{c|}{\textbf{\begin{tabular}[c]{@{}c@{}}Commu-\\ nication\end{tabular}}} & \textbf{Accuracy} & \textbf{\begin{tabular}[c]{@{}c@{}}Comp-\\ leteness\end{tabular}} & \textbf{\begin{tabular}[c]{@{}c@{}}Context\\ Awareness\end{tabular}} \\ \midrule
\multicolumn{9}{c}{\cellcolor[HTML]{EFEFEF}Medical Specialized Models} \\
UltraMedical-70B & \multicolumn{1}{c|}{{\color[HTML]{000000} 23.45}} & {\color[HTML]{000000} 17.70} & {\color[HTML]{000000} 30.77} & {\color[HTML]{000000} 20.39} & \multicolumn{1}{c|}{{\color[HTML]{000000} 20.28}} & {\color[HTML]{000000} 29.54} & {\color[HTML]{000000} 25.03} & {\color[HTML]{000000} 30.99} \\
Llama3-Med42-70B & \multicolumn{1}{c|}{{\color[HTML]{000000} 24.45}} & {\color[HTML]{000000} 15.19} & {\color[HTML]{000000} 28.97} & {\color[HTML]{000000} 22.48} & \multicolumn{1}{c|}{{\color[HTML]{000000} 32.43}} & {\color[HTML]{000000} 33.87} & {\color[HTML]{000000} 25.78} & {\color[HTML]{000000} 31.20} \\
MedCritical-7B & \multicolumn{1}{c|}{24.73} & 20.04 & 35.33 & 25.52 & \multicolumn{1}{c|}{22.20} & 33.05 & 31.73 & 32.03 \\
Baichuan-M2-32B & \multicolumn{1}{c|}{38.76} & 35.23 & 41.47 & 33.38 & \multicolumn{1}{c|}{46.11} & 36.10 & 37.55 & 34.04 \\ \midrule
\multicolumn{9}{c}{\cellcolor[HTML]{EFEFEF}Open-Source Models} \\
Qwen3-235B-A22B & \multicolumn{1}{c|}{40.24} & 32.59 & 43.44 & 34.91 & \multicolumn{1}{c|}{50.30} & 40.10 & 37.51 & 31.42 \\
Kimi-K2 & \multicolumn{1}{c|}{42.71} & 33.38 & 41.79 & 38.73 & \multicolumn{1}{c|}{52.50} & 41.56 & 41.10 & 33.29 \\
DeepSeek-R1 & \multicolumn{1}{c|}{42.94} & 39.53 & 48.85 & 39.02 & \multicolumn{1}{c|}{48.16} & 41.89 & 47.29 & 31.97 \\ \midrule
\multicolumn{9}{c}{\cellcolor[HTML]{EFEFEF}Proprietary Models} \\
GPT-4o & \multicolumn{1}{c|}{19.35} & 12.49 & 20.22 & 19.57 & \multicolumn{1}{c|}{24.80} & 25.92 & 29.13 & 26.68 \\
OpenAI-o1 & \multicolumn{1}{c|}{32.03} & 21.14 & 33.77 & 26.85 & \multicolumn{1}{c|}{40.25} & 34.95 & 29.89 & 30.92 \\
GPT-4.1 & \multicolumn{1}{c|}{34.48} & 21.65 & 33.24 & 27.87 & \multicolumn{1}{c|}{45.92} & 36.66 & 31.42 & 30.03 \\
OpenAI-o4-mini & \multicolumn{1}{c|}{39.46} & 32.25 & 40.85 & 35.61 & \multicolumn{1}{c|}{52.11} & 40.78 & 44.19 & 31.14 \\
Grok-3 & \multicolumn{1}{c|}{42.51} & 34.16 & 41.97 & 37.83 & \multicolumn{1}{c|}{54.97} & 41.07 & 38.84 & \textbf{37.62} \\
Gemeni-2.5-Pro & \multicolumn{1}{c|}{43.49} & 34.42 & 44.48 & 38.85 & \multicolumn{1}{c|}{51.46} & 41.32 & 39.49 & 34.08 \\
Claude-4-Sonnet & \multicolumn{1}{c|}{44.28} & 35.10 & 46.50 & 40.91 & \multicolumn{1}{c|}{50.64} & 40.63 & 40.86 & 36.26 \\ \midrule
\textbf{TheraAgent(Ours)} & \multicolumn{1}{c|}{\textbf{48.94}} & \textbf{47.49} & \textbf{55.63} & \textbf{44.65} & \multicolumn{1}{c|}{\textbf{55.29}} & \textbf{44.80} & \textbf{51.72} & 37.16 \\ \bottomrule
\end{tabular}%
}
\vspace{-0.5em}
\caption{Performance comparison on HealthBench across different models. We report the overall score, theme-level and axis-level scores for medical, open-source, and proprietary models. The best results are bolded.}
\label{tab:main_result}
\end{table*}

\subsection{Memorizer}
The \textbf{Memorizer} maintains a structured repository of historical treatment plans together with their corresponding evaluations, enabling experience accumulation across iterations. Each memory item $\mathbf{M}_i = (\mathbf{T}_i, \mathbf{R}_i, \mathbf{s}_i)$ encapsulates the generated schedule, reasoning traces, and evaluation scores from iteration $i$.
At iteration $k$, the memory state is defined as the collection of all past memory items as $\mathcal{M}^{(k-1)} = \{\mathbf{M}_i\}_{i=1}^{k-1}$, which is incrementally updated as new treatment plans are generated and evaluated.

To facilitate refinement, the Planner performs score-aware retrieval, selecting a subset of memory items with the $Top\text{-}N$highest scores in  $\mathcal{M}^{(k-1)}$. By prioritizing high-quality historical schedules, the Memorizer stabilizes the self-improving process, enabling TheraAgent to progressively improve therapeutic quality with reduced computational cost.

\subsection{Agent Output}
TheraAgent outputs the schedule $\mathbf{T}^{*}$ by maximizing the score in the final $L$ iterations shown as:
\begin{equation}
\mathbf{T}^{*} = \underset{\mathbf{T}_k, k \in \{N-L+1, \dots, N\}}{\arg\max} s_k.
\end{equation}
This selection leverages the performance gains from accumulated feedback while mitigating potential late-stage fluctuations.
To optimize efficiency, an early stopping mechanism terminates the process if scores consistently meet a threshold $\tau$ for three consecutive iterations formed as $s_{(k-2)}, s_{(k-1)}, s_k \geq \tau$. This method ensures high-quality stability while minimizing unnecessary computational overhead.

In summary, through the integration of iterative generation, multi-dimensional judging and memory-guided refinement, TheraAgent enhances the accuracy, completeness, and safety of treatment plans.

\section{Experiments}

\subsection{Dataset}

\paragraph{HealthBench.} 

HealthBench provides a comprehensive and realistic benchmark for evaluating real-world healthcare capabilities~\cite{arora2025healthbenchevaluatinglargelanguage}. As our study focuses on the task of treatment planning, we first filter the original dataset to retain only treatment-related conversations using GPT-4.1 as a classifier. The filtered samples are categorized into four medical departments: endocrinology (265), gastroenterology (262), neurology (395), and respiratory (319), totaling 1,241 cases for evaluation. Detailed dataset statistics and department-wise distributions are provided in Appendix~\ref{app:healthbench dataset}.

\paragraph{Real-World Case Dataset.}

We incorporate 35 physician-authored cases from a public platform\footnote{https://www.yiigle.com/Paper/}, with 9 cases each from respiratory, neurology, cardiology, and 8 cases from ophthalmology. These cases represent real-world, clinical scenarios with complex therapeutic challenges. Additional case descriptions are provided in Appendix~\ref{app:real-world case dataset}.

\begin{figure*}[t]
\centering
\includegraphics[width=\textwidth]{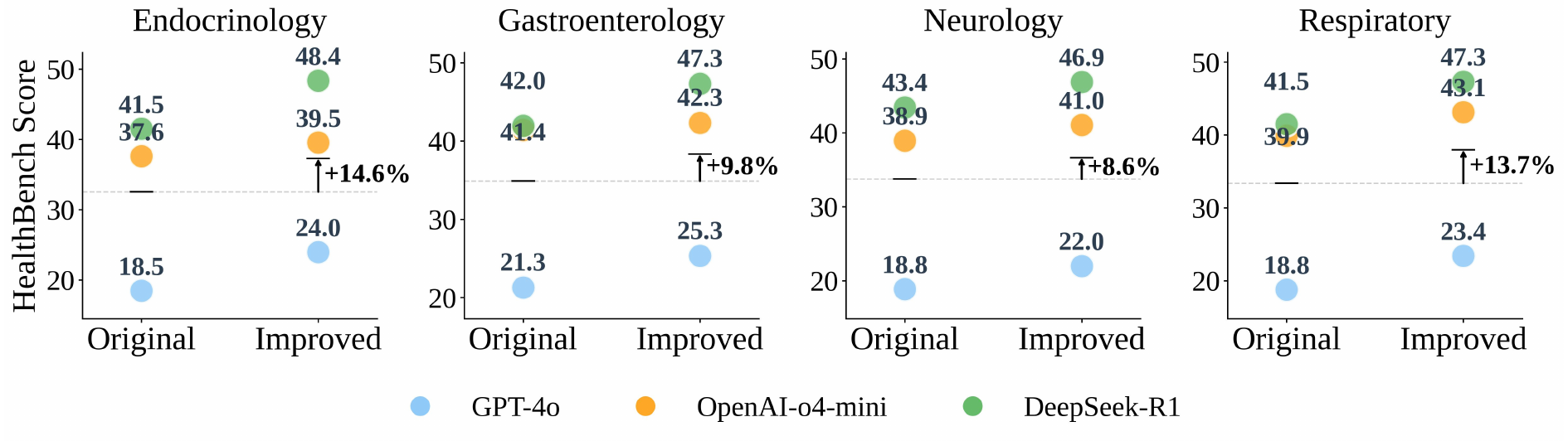}
\vspace{-2em}
\caption{Generalization analysis of TheraAgent across four medical departments. The plot compares the HealthBench scores of three backbone models in their "Original" (base) state versus the "Improved" state (using TheraAgent framework). Horizontal bold lines represent the mean score of the three models in each state. The annotated values indicate the relative improvement rate, calculated as the increase in the mean score divided by the original mean.}
\label{fig:generalization}
\vspace{-0.5em}
\end{figure*}

\subsection{Evaluation Metrics}

\paragraph{Automatic evaluation.}
We adopt HealthBench as the automatic evaluation for large-scale experiments. Following its specific rubric, responses are assessed across several medical dimensions. The hand-crafted rubrics comprehensively capture the real-world expectations for a treatment plan. GPT-4.1 is used as the rubric evaluator to ensure reliable and efficient scoring, consistent with prior HealthBench evaluations. 

\paragraph{Medical Experts annotation.}
We further conduct human evaluation on the Real-World Case dataset with licensed medical experts who have over four years of clinical experience. Experts compare treatment plans generated by TheraAgent, raw LLMs, and physician-authored references. Detailed annotation protocols, evaluation criteria, and interface designs are provided in Appendix~\ref{app:human annotation}.

\subsection{Baselines and Hyperparameters}

\paragraph{Baselines.}
We compare TheraAgent against a representative set of strong baselines. Specifically, we include MedCritical~\cite{su2025medcritical} and Baichuan-M2~\cite{m2team2025baichuanm2scalingmedicalcapability} as medical-specialized models; DeepSeek-R1~\cite{deepseekai2025deepseekr1incentivizingreasoningcapability} and Kimi-K2~\cite{kimiteam2025kimik2openagentic} as open-source general models; and Grok-3~\cite{grok3_2025}, and Claude-4-Sonnet~\cite{claude4_system_card_2025} as proprietary baselines. Closed-source agentic methods such as MedPlan~\cite{hsu2025medplan} are excluded due to limited accessibility. A complete list of evaluated models and implementation details are provided in Appendix~\ref{app:experiment_details}. 

\paragraph{TheraAgent configuration.}
TheraAgent utilizes DeepSeek-R1 as the backbone for both the Planner and TheraJudge. In each iteration, the Planner retrieves $Top\text{-}N=3$ memory items. The early stopping threshold is set to $\tau = 98$ with a maximum of 10 iterations, while the output window $L=3$. To stabilize evaluation, TheraJudge incorporates $3$ few-shot exemplars per department. Notably, RAG is disabled during HealthBench evaluation to mitigate potential biases from region-specific guidelines, ensuring a fair assessment grounded in universal clinical reasoning. To ensure experimental stability, advanced models including Qwen-3-235B-A22B, Kimi-K2, DeepSeek-R1, Gemini-2.5-Pro, Claude-4-Sonnet and TheraAgent are evaluated based on the average performance across three runs.


\begin{figure*}[h]
\centering
\includegraphics[width=\textwidth]{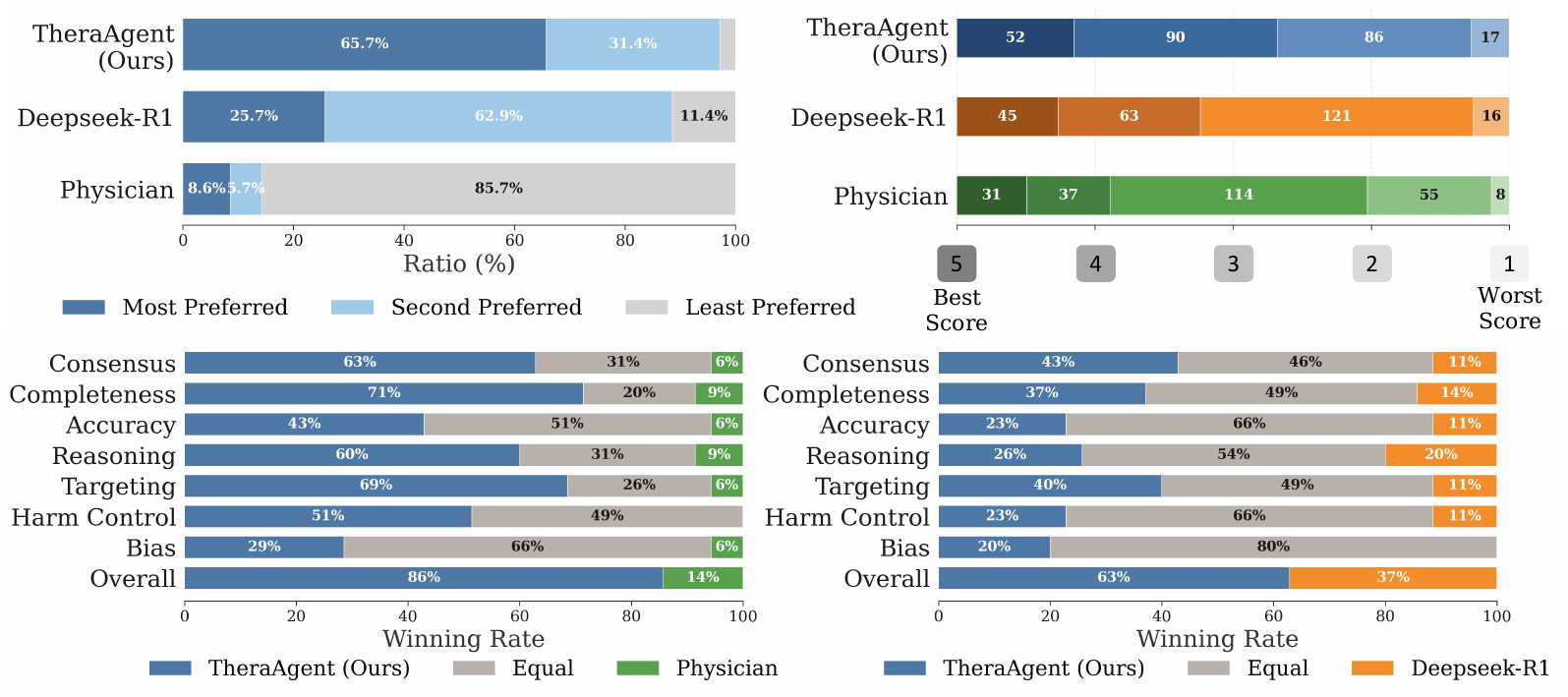}
\vspace{-1.5em}
\caption{\textbf{Expert evaluation on Real Medical Cases.} Top: Three-way preference rankings (left) and 5-point rating distributions (right), with numbers indicating the absolute count for each score. Bottom: Pairwise comparisons across seven clinical dimensions against human physicians (left) and DeepSeek-R1 (right). 
}
\label{fig:expert annotation}
\vspace{-0.5em}
\end{figure*}

\subsection{Main Results}

\paragraph{Automatic evaluation results.}The performance of TheraAgent on the HealthBench subset is summarized in Table~\ref{tab:main_result}. Overall, TheraAgent achieves a state-of-the-art score of 48.94, surpassing the second-best model by 4.66 points.
A dimension-wise analysis further shows that TheraAgent outperforms all baselines across several critical dimensions. Notably, it achieves substantial improvements in \textit{Hedging} of 6.78 points and \textit{Completeness} of 4.43 points, indicating more precise clinical decision-making and reduced omissions in treatment planning. In addition, TheraAgent attains a state-of-the-art score of 44.65 in \textit{Context Seeking}, reflecting its enhanced ability to actively elicit and integrate specific information for more targeted therapeutic planning. Complete results across all evaluated models are reported in Appendix~\ref{app:all results of healthbench}.

To assess robustness across models, we perform a stratified analysis over different models. As illustrated in Figure~\ref{fig:generalization}, TheraAgent consistently enhances the treatment planning capabilities of diverse backbones, ranging from standard models like GPT-4o to reasoning-specialized models like DeepSeek-R1. Across four distinct medical departments, the framework yields uniform performance boosts, with average relative improvements ranging from \textbf{+8.6\%} to \textbf{+14.6\%}. Notably, every model in every department exhibits a positive trajectory from its "Original" to "Improved" state.

\paragraph{Human annotation results.} To assess clinical utility, we conducted a comprehensive blinded evaluation on the Real-World Case dataset involving licensed experts, comparing TheraAgent against both human physicians and the base model (DeepSeek-R1). 
As illustrated in Figure~\ref{fig:expert annotation} (top-left), TheraAgent demonstrates a dominant superiority in three-way preference rankings, being selected as the “Most Preferred” in \textbf{65.7\%} of cases, substantially outperforming both the one-shot base model in 25.7\% and human physicians in 8.6\%. We further analyze the factors underlying the comparatively lower performance of human physicians in Section~\ref{sec:case_study}.

Detailed pairwise comparisons (Figure~\ref{fig:expert annotation}, bottom) further reveal the specific drivers of superiority of TheraAgent. Compared to the base model (DeepSeek-R1), TheraAgent achieves a 63\% overall win rate, with the gains in \textit{Targeting} of 40\% and \textit{Consensus} of 43\%, indicating more patient-specific and precise treatment plans. The detailed dimension-wise rating results are provided in Appendix~\ref{app:rating_radar}, which further corroborate the observed performance advantages of TheraAgent.
Overall, these results demonstrate that TheraAgent’s self-improving pipeline effectively produces precise, safe, and comprehensive treatment regimens.

\begin{table}[t]
\centering
\resizebox{\columnwidth}{!}{%
\begin{tabular}{lrrr}
\hline
Evaluation & Spearman & Pearson & CCC \\ \hline
BLEU & 0.0000 & -0.0221 & -0.0179 \\
ROUGE1 & 0.2052 & 0.2179 & 0.1989 \\
ROUGE2 & 0.1118 & 0.1194 & 0.1075 \\
ROUGEL & 0.1118 & 0.1516 & 0.1348 \\
BERT Score & 0.1539 & 0.1106 & 0.1009 \\
LLM Score & 0.1539 & 0.1089 & 0.0879 \\
\rowcolor{gray!20}
\textbf{TheraJudge (Ours)} & \textbf{0.6669} & \textbf{0.7052} & \textbf{0.6467} \\ \hline
\end{tabular}%
}
\caption{\textbf{Consistency experiment of different evaluation methods against HealthBench evaluation.} The correlation between distinct method scores and HealthBench overall scores, are evaluated across five model outputs per case. Reported values are the median scores across all cases of neurology department.}
\label{tab:healthbench_judge}
\vspace{-0.5em}

\end{table}
\paragraph{Judge Agreement with HealthBench}
To validate the reliability of our internal critic, we assess the alignment between TheraJudge and official HealthBench evaluation (Table~\ref{tab:healthbench_judge}), following the experimental settings described in Appendix~\ref{app:judge_healthbench_setup}. TheraJudge demonstrates high consistency with established standards, achieving a Pearson correlation of 0.7052 and Spearman correlation of 0.6669. In contrast, traditional lexical metrics (e.g., BLEU, ROUGE) and vanilla LLM scoring exhibit weak correlations, highlighting their inability to capture the complex semantic nuances of clinical reasoning. This alignment is critical for the success of the TheraAgent framework. It indicates that \textbf{\textit{TheraJudge acts as a faithful proxy for expert evaluation}}, providing valid feedback signals regarding safety and completeness rather than arbitrary noise.


\section{Analysis}
\subsection{Inference-Time Scaling}
\begin{figure}[t]
\centering
\includegraphics[width=\columnwidth]{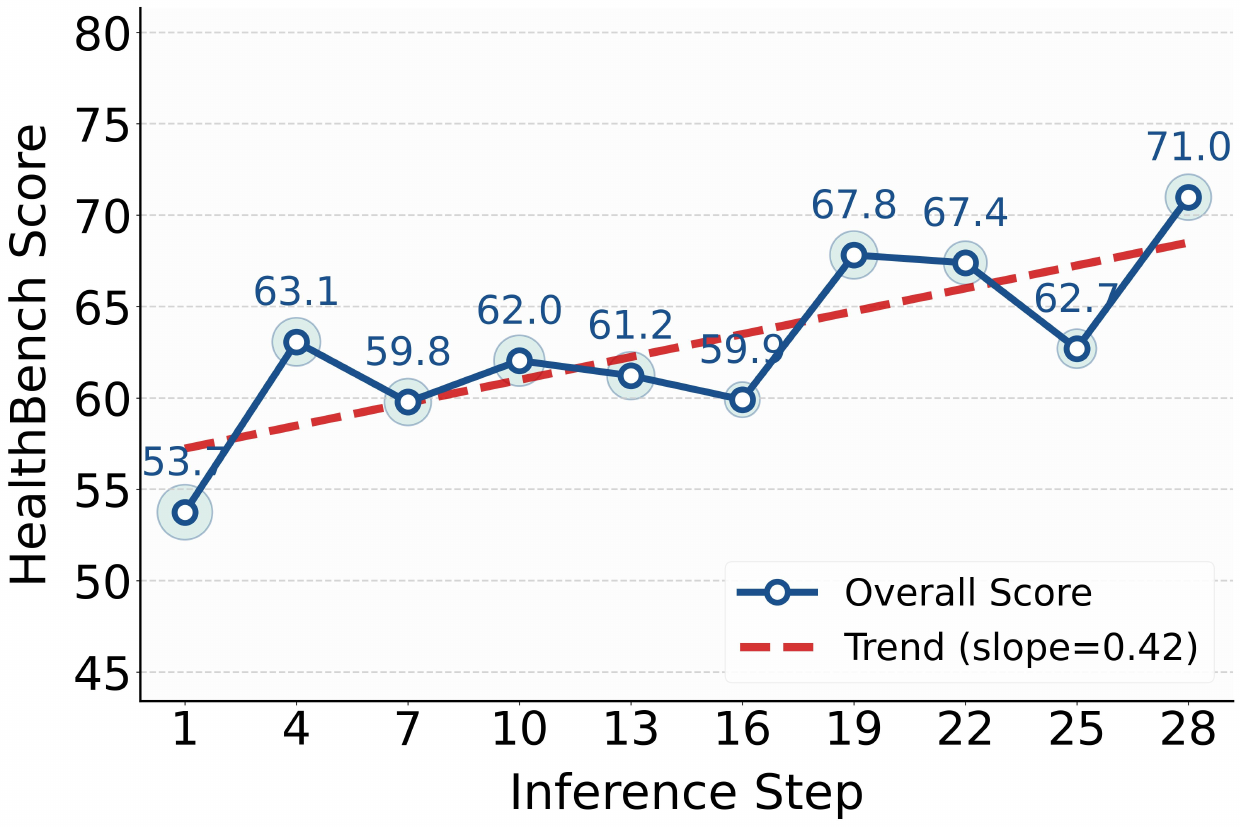}
\vspace{-2em}
\caption{\textbf{Inference-time scaling in TheraAgent:} performance progressively improves across inference steps. Each point denotes the mean HealthBench score over cases, and the red dashed line ( \textcolor{red}{\textbf{- -}} ) shows an overall positive performance trend cross iterations.}
\label{fig:iterative}
\end{figure}

We conduct inference time experiments on 10 cases of Neurology due to the high cost of HealthBench evaluation at each step. Figure~\ref{fig:iterative} illustrates the evolution of HealthBench score as the number of inference steps increases. As shown in the figure, the scores exhibit a upward trend over successive iterations. The fitted linear trend further indicates a positive correlation between iteration count and evaluation quality. This behavior indicates that the self-improving framework effectively leverages feedback to refine subsequent generations.

\subsection{Cost and Efficiency Analysis}

\begin{table}[t]
\centering

\resizebox{\columnwidth}{!}{%
\begin{tabular}{lcccc}
\toprule
\textbf{Method} & \textbf{Calls} & \textbf{Tokens} & \textbf{Time (s)} & \textbf{Relative Cost} \\
\midrule
DeepSeek-R1       & 1  & 1,358  & 30.6  & 1.0$\times$  \\
Kimi-K2           & 1  & 1,764  & 16.2  & 2.1$\times$  \\
Claude-4-Sonnet   & 1  & 1,295  & 23.6  & 6.2$\times$  \\
Gemini-2.5-Pro    & 1  & 3,925  & 50.4  & 11.1$\times$ \\
\midrule
TheraAgent (3 iter)  & 6  & 13,445 & 332.6 & 9.9$\times$  \\
TheraAgent (10 iter) & 20 & 87,005 & 753.5 & 64.1$\times$ \\
\bottomrule
\end{tabular}%
}
\caption{\textbf{Computational efficiency and relative cost comparison.} Statistics are averaged over three representative cases. Relative cost is normalized against DeepSeek-R1.}
\label{tab:cost}
\end{table}

While effective, iterative refinement inherently introduces higher computational overhead than single-pass baselines. As shown in Table~\ref{tab:cost}, a single call to DeepSeek-R1 requires 1,358 tokens and 30.6 seconds, whereas TheraAgent operating at 3 iterations uses 6 API calls, 13,445 tokens and 332.6 seconds (9.9x relative cost). Extending to 10 iterations further increases usage to 20 calls and 87,005 tokens (64.1x relative cost).

This trade-off is intentional, as it enables test-time scaling in high-stakes treatment planning, where thorough guideline adherence and safety are paramount and justify the additional cost. Furthermore, the resulting high-quality iterative reasoning trajectories could be distilled into more efficient model parameters via supervised fine-tuning or reinforcement learning, potentially achieving comparable reliability with reduced inference latency in future deployments.

\subsection{Generalization on Additional Benchmark}

\begin{table}[t]
\centering

\resizebox{\columnwidth}{!}{%
\begin{tabular}{lccccc}
\toprule
\textbf{Model} & \textbf{Overall} & \textbf{\begin{tabular}[c]{@{}c@{}}Neur- \\ology\end{tabular}} & \textbf{\begin{tabular}[c]{@{}c@{}}Respi- \\ratory\end{tabular}} & \textbf{\begin{tabular}[c]{@{}c@{}}Endo- \\crinology\end{tabular}} & \textbf{\begin{tabular}[c]{@{}c@{}}Gastro- \\enterology\end{tabular}} \\
\midrule
Kimi-K2          & 55.72 & 55.20 & 59.76 & 55.20 & 49.33 \\
DeepSeek-R1      & 57.60 & 51.20 & 62.00 & 60.27 & 50.00 \\
Gemini-2.5-Pro   & 61.63 & 66.40 & 65.45 & 64.93 & 52.79 \\
Claude-4-Sonnet  & 63.92 & 63.20 & 67.68 & 63.20 & \textbf{58.13} \\
\midrule
TheraAgent       & \textbf{69.32} & \textbf{75.20} & \textbf{73.44} & \textbf{76.27} & 58.00 \\
\bottomrule
\end{tabular}%
}
\caption{\textbf{Extended evaluation on the MTMedDialog Benchmark}, an open-domain medical dialogue dataset focused on diagnostic and therapeutic tasks}
\label{tab:mtmeddialog_extended}
\vspace{-0.5em}
\end{table}

To demonstrate that the capabilities of TheraAgent extend beyond a single evaluation framework, we conducted additional experiments on MTMedDialog~\cite{feng2025doctoragentrlmultiagentcollaborativereinforcement}, an open-domain medical dialogue dataset focused on diagnostic and therapeutic tasks. We sample 15 cases per department and adopt the same evaluation protocol.

As shown in Table~\ref{tab:mtmeddialog_extended}, TheraAgent achieves the best overall score of 79.67, outperforming strong baselines including Claude-4-Sonnet (78.33), Gemini-2.5-Pro (76.55), and Kimi-K2 (75.00). The improvements are particularly notable in the Neurology and Respiratory departments, where TheraAgent scores 86.67 and 77.33, respectively. These findings are consistent with our primary experiments on HealthBench, confirming that the performance gains are not benchmark-specific, but generalize effectively across datasets with different case distributions and dialogue structures.

\subsection{Component Ablation}

\begin{table}[t]
\centering

\resizebox{\columnwidth}{!}{%
\begin{tabular}{cccc}
\hline
Dimensions & Fewshots & RAG & \multicolumn{1}{c}{HealthBench Score} \\ \hline
\checkmark &  &  & 48.66 \\
 & \checkmark &  & 50.62 \\
  &  & \checkmark & 45.98 \\
\checkmark & \checkmark &  & \textbf{52.36} \\
\checkmark & \checkmark & \checkmark & 45.96 \\ \hline
\multicolumn{3}{c}{Base Model w/o Judge} & 41.15 \\
\multicolumn{3}{c}{Vanilla Judge} & 48.50 \\
\hline
\end{tabular}%
}
\caption{Ablation study of TheraJudge components.}
\label{Judge_ablation}
\vspace{-12pt}

\end{table}
Table~\ref{Judge_ablation} presents an ablation study of the three components of TheraJudge. All variants outperform the base model without a judge, which scores 41.15, underscoring the importance of structured evaluation. Among individual components, few-shot exemplars yield the strongest improvement with a score of 50.62, indicating their effectiveness in stabilizing evaluation behavior, while dimensional scoring also brings consistent gains with a score of 48.66. In contrast, RAG alone results in relatively modest improvement at 45.98, likely due to guideline heterogeneity in the globally sourced HealthBench dataset. The best performance is obtained by combining few-shot exemplars with dimensional scoring, reaching a score of 52.36. Incorporating RAG further reduces performance to 45.96, which supports our decision to disable RAG during HealthBench evaluation. Additional ablation results on the Memorizer and RAG components are provided in Appendix~\ref{app:ablations}.


\subsection{Case Study}
\label{sec:case_study}
A comparison of treatment plans for a CPFE case (Appendix~\ref{app:case study}) exhibits TheraAgent's clinical superiority across two key dimensions.
\textbf{Information Accuracy:} TheraAgent avoids generic errors by tailoring decisions to specific patient data. It correctly withheld antibiotics based on the patient's normal inflammatory markers and explicitly conditioned antifibrotic therapy on HRCT findings, ensuring interventions were grounded in the individual's diagnostic context.
\textbf{Harm Control:} DeepSeek-R1 exhibits critical errors, including subtherapeutic N-acetylcysteine dosing (600mg/day vs required 1,800mg/day) and premature morphine use. TheraAgent ensures safety by correcting these dosages, implementing strict intervention thresholds, and mandating continuous monitoring of respiratory and renal functions.

In this case, the physician’s plan explains its lower expert annotation scores by focusing on core antifibrotic therapy and general monitoring while omitting explicit decision criteria (e.g., HRCT pattern–dependent treatment). Such condensation is common in routine clinical practice due to time, cost, and workflow constraints. TheraAgent bridges this gap by converting implicit clinical logic into explicit requirements. In deployment, TheraAgent suggests structured draft plans that physicians can selectively adopt or modify, and acts as a safety reminder by highlighting thresholds, contraindications, and monitoring needs, thereby reducing oversight in complex cases.

\section{Conclusion}
In this work, we redefine treatment planning not as simple generation, but as a distinct reasoning and refinement task. We propose TheraAgent, an agentic framework that replaces one-shot generation with a \textit{generate–reflect–refine} pipeline, mimicking the iterative cognitive process of clinicians. Central to this approach is TheraJudge, a clinically aligned internal critic that assesses plans against key dimensions like accuracy and safety, driving active self-correction.
Experiments on HealthBench and expert-annotated real-world cases demonstrate that TheraAgent achieves state-of-the-art performance and an 86\% win rate against human physicians. Furthermore, TheraJudge exhibits high agreement with HealthBench evaluation, validating its dual role as both an optimization signal and a reliable metric. Our findings show that integrating domain-specific evaluation into the inference loop is essential for transforming LLMs into safe, precise, and practical clinical planners.

\section*{Acknowledgements}
This work was partly supported by the Fundamental and Interdisciplinary Disciplines Breakthrough Plan of the Ministry of Education of China (No. JYB2025XDXM101), the National Natural Science Foundation of China (62372260, 62276152), and Wuxi Research Institute of Applied Technologies Tsinghua University. Weizhi Ma was also supported by the Beijing Nova Program.

\section*{Limitations}

There are some limitations in our work. 
First, our experimental validation of the iterative framework was primarily conducted using specific high-performance backbone models (DeepSeek-R1, GPT-4o, and OpenAI-o4-mini). While these models demonstrate the effectiveness of the approach, the generalizability of TheraAgent across a broader spectrum of smaller-scale models remains to be fully characterized. 
Second, the self-improving design of TheraAgent introduces higher computational overhead than single-pass generation. Although we employ score-aware memory retrieval and early-stopping mechanisms to improve efficiency, the framework may still be less suitable for real-time or resource-constrained clinical settings without further optimization. 
Third, TheraAgent focuses on text-based treatment planning and does not directly incorporate structured clinical signals such as laboratory time series, imaging data, or real-time patient monitoring. Extending the framework to multimodal clinical inputs remains an important direction for future work. 

\section*{Ethical Considerations}

The application of large language models to clinical decision support raises important ethical and safety considerations. TheraAgent is designed to assist treatment planning rather than to replace professional medical judgment, and its outputs should not be treated as definitive clinical decisions. Despite its strong empirical performance, the framework may still produce incomplete, incorrect, or context-insensitive recommendations, particularly in the presence of rare conditions, atypical patient profiles, or missing clinical information. As such, its deployment requires appropriate clinical oversight, clear usage boundaries, and integration into existing medical workflows with human-in-the-loop safeguards.

To mitigate potential risks, TheraAgent emphasizes transparency through explicit reasoning traces, structured feedback, and guideline-grounded evaluation. These mechanisms enable pyhsicians to interpret, verify, and critique generated treatment plans, reducing the risk of uncritical reliance on model outputs.

Regarding data ethics and privacy, our study relies solely on publicly available or authorized datasets, with all patient cases fully de-identified. We further acknowledge that clinical guidelines and standards of care vary across regions. Accordingly, we controlled the use of external clinical guidelines to ensure that TheraAgent does not blindly rely on any specific regional or source guideline, respecting variations across different regions. Future deployments should incorporate localized clinical oversight and regulatory compliance to ensure responsible real-world use.

\bibliography{custom}

@article{Evaluation_of_large,
   author = {Chase, Aaron and Most, Amoreena and Sikora, Andrea and Smith, Susan E. and Devlin, John W. and Xu, Shaochen and Liu, Tianming and Murray, Brian},
   title = {Evaluation of large language models' ability to identify clinically relevant drug-drug interactions and generate high-quality clinical pharmacotherapy recommendations},
   journal = {American Journal of Health-System Pharmacy},
   ISSN = {1079-2082},
   DOI = {10.1093/ajhp/zxaf168},
   url = {<Go to ISI>://WOS:001530900400001},
   year = {2025},
   type = {Journal Article}
}

@article{exploring_the_role,
   author = {Hassan, Bassam Abdul Rasool and Mohammed, Ali Haider and Hallit, Souheil and Malaeb, Diana and Hosseini, Hassan},
   title = {Exploring the role of artificial intelligence in chemotherapy development, cancer diagnosis, and treatment: present achievements and future outlook},
   journal = {Frontiers in Oncology},
   volume = {15},
   ISSN = {2234-943X},
   DOI = {10.3389/fonc.2025.1475893},
   url = {<Go to ISI>://WOS:001427980600001},
   year = {2025},
   type = {Journal Article}
}

@article{ai_driven_clinical,
   author = {Khude, Hrishikesh and Shende, Pravin},
   title = {{AI}-driven clinical decision support systems: Revolutionizing medication selection and personalized drug therapy},
   journal = {Advances in Integrative Medicine},
   volume = {12},
   number = {4},
   ISSN = {2212-9588},
   DOI = {10.1016/j.aimed.2025.100529},
   url = {<Go to ISI>://WOS:001525455200001},
   year = {2025},
   type = {Journal Article}
}

@article{performance_analysis_of,
   author = {Lin, Zhiwu and Li, Yuanyuan and Wu, Min and Liu, Hongmei and Song, Xiaoyang and Yu, Qian and Xiao, Guibao and Xie, Jiajun},
   title = {Performance analysis of large language models {Chatgpt-4o}, {OpenAI O1}, and {OpenAI O3} mini in clinical treatment of pneumonia: a comparative study},
   journal = {Clinical and Experimental Medicine},
   volume = {25},
   number = {1},
   ISSN = {1591-8890},
   DOI = {10.1007/s10238-025-01743-7},
   url = {<Go to ISI>://WOS:001512309800001},
   year = {2025},
   type = {Journal Article}
}

@article{effectiveness_of_chat,
   author = {Roosan, Don and Padua, Pauline and Khan, Raiyan and Khan, Hasiba and Verzosa, Claudia and Wu, Yanting},
   title = {Effectiveness of {ChatGPT} in clinical pharmacy and the role of artificial intelligence in medication therapy management},
   journal = {Journal of the American Pharmacists Association},
   volume = {64},
   number = {2},
   ISSN = {1544-3191},
   DOI = {10.1016/j.japh.2023.11.023},
   url = {<Go to ISI>://WOS:001225862200001},
   year = {2024},
   type = {Journal Article}
}

@article{a_generalist_medical,
   author = {Liu, Xiaohong and Liu, Hao and Yang, Guoxing and Jiang, Zeyu and Cui, Shuguang and Zhang, Zhaoze and Wang, Huan and Tao, Liyuan and Sun, Yongchang and Song, Zhu and Hong, Tianpei and Yang, Jin and Gao, Tianrun and Zhang, Jiangjiang and Li, Xiaohu and Zhang, Jing and Sang, Ye and Yang, Zhao and Xue, Kanmin and Wu, Song and Zhang, Ping and Yang, Jian and Song, Chunli and Wang, Guangyu},
   title = {A generalist medical language model for disease diagnosis assistance},
   journal = {Nature Medicine},
   volume = {31},
   number = {3},
   ISSN = {1078-8956},
   DOI = {10.1038/s41591-024-03416-6},
   url = {<Go to ISI>://WOS:001391696700001
https://www.nature.com/articles/s41591-024-03416-6.pdf},
   year = {2025},
   type = {Journal Article}
}

@article{toward_accurate_differential,
   author = {McDuff, Daniel and Schaekermann, Mike and Tu, Tao and Palepu, Anil and Wang, Amy and Garrison, Jake and Singhal, Karan and Sharma, Yash and Azizi, Shekoofeh and Kulkarni, Kavita and Hou, Le and Cheng, Yong and Liu, Yun and Mahdavi, S. Sara and Prakash, Sushant and Pathak, Anupam and Semturs, Christopher and Patel, Shwetak and Webster, Dale R. and Dominowska, Ewa and Gottweis, Juraj and Barral, Joelle and Chou, Katherine and Corrado, Greg S. and Matias, Yossi and Sunshine, Jake and Karthikesalingam, Alan and Natarajan, Vivek},
   title = {Towards accurate differential diagnosis with large language models},
   journal = {Nature},
   volume = {642},
   number = {8067},
   ISSN = {0028-0836},
   DOI = {10.1038/s41586-025-08869-4},
   url = {<Go to ISI>://WOS:001463961800001
https://www.nature.com/articles/s41586-025-08869-4.pdf},
   year = {2025},
   type = {Journal Article}
}

@article{toward_expert_level,
   author = {Singhal, Karan and Tu, Tao and Gottweis, Juraj and Sayres, Rory and Wulczyn, Ellery and Amin, Mohamed and Hou, Le and Clark, Kevin and Pfohl, Stephen R. and Cole-Lewis, Heather and Neal, Darlene and Rashid, Qazi Mamunur and Schaekermann, Mike and Wang, Amy and Dash, Dev and Chen, Jonathan H. and Shah, Nigam H. and Lachgar, Sami and Mansfield, Philip Andrew and Prakash, Sushant and Green, Bradley and Dominowska, Ewa and Aguera y Arcas, Blaise and Tomasev, Nenad and Liu, Yun and Wong, Renee and Semturs, Christopher and Mahdavi, S. Sara and Barral, Joelle K. and Webster, Dale R. and Corrado, Greg S. and Matias, Yossi and Azizi, Shekoofeh and Karthikesalingam, Alan and Natarajan, Vivek},
   title = {Toward expert-level medical question answering with large language models},
   journal = {Nature Medicine},
   volume = {31},
   number = {3},
   ISSN = {1078-8956},
   DOI = {10.1038/s41591-024-03423-7},
   url = {<Go to ISI>://WOS:001393036000001
https://www.nature.com/articles/s41591-024-03423-7.pdf},
   year = {2025},
   type = {Journal Article}
}

@article{towards_conversational_diag,
   author = {Tu, Tao and Schaekermann, Mike and Palepu, Anil and Saab, Khaled and Freyberg, Jan and Tanno, Ryutaro and Wang, Amy and Li, Brenna and Amin, Mohamed and Cheng, Yong and Vedadi, Elahe and Tomasev, Nenad and Azizi, Shekoofeh and Singhal, Karan and Hou, Le and Webson, Albert and Kulkarni, Kavita and Mahdavi, S. Sara and Semturs, Christopher and Gottweis, Juraj and Barral, Joelle and Chou, Katherine and Corrado, Greg S. and Matias, Yossi and Karthikesalingam, Alan and Natarajan, Vivek},
   title = {Towards conversational diagnostic artificial intelligence},
   journal = {Nature},
   volume = {642},
   number = {8067},
   ISSN = {0028-0836},
   DOI = {10.1038/s41586-025-08866-7},
   url = {<Go to ISI>://WOS:001462553800001
https://www.nature.com/articles/s41586-025-08866-7.pdf},
   year = {2025},
   type = {Journal Article}
}

@inproceedings{diao2025guidebench,
    title = "{G}uide{B}ench: Benchmarking Domain-Oriented Guideline Following for {LLM} Agents",
    author = "Diao, Lingxiao  and
      Xu, Xinyue  and
      Sun, Wanxuan  and
      Yang, Cheng  and
      Zhang, Zhuosheng",
    editor = "Che, Wanxiang  and
      Nabende, Joyce  and
      Shutova, Ekaterina  and
      Pilehvar, Mohammad Taher",
    booktitle = "Proceedings of the 63rd Annual Meeting of the Association for Computational Linguistics (Volume 1: Long Papers)",
    month = jul,
    year = "2025",
    address = "Vienna, Austria",
    publisher = "Association for Computational Linguistics",
    url = "https://aclanthology.org/2025.acl-long.557/",
    doi = "10.18653/v1/2025.acl-long.557",
    pages = "11361--11399",
    ISBN = "979-8-89176-251-0"
}

@inproceedings{huang2024queryagent,
    title = "{Q}uery{A}gent: A Reliable and Efficient Reasoning Framework with Environmental Feedback based Self-Correction",
    author = "Huang, Xiang  and
      Cheng, Sitao  and
      Huang, Shanshan  and
      Shen, Jiayu  and
      Xu, Yong  and
      Zhang, Chaoyun  and
      Qu, Yuzhong",
    editor = "Ku, Lun-Wei  and
      Martins, Andre  and
      Srikumar, Vivek",
    booktitle = "Proceedings of the 62nd Annual Meeting of the Association for Computational Linguistics (Volume 1: Long Papers)",
    month = aug,
    year = "2024",
    address = "Bangkok, Thailand",
    publisher = "Association for Computational Linguistics",
    url = "https://aclanthology.org/2024.acl-long.274/",
    doi = "10.18653/v1/2024.acl-long.274",
    pages = "5014--5035"
}

@inproceedings{liu2025cer,
    title = "Contextual Experience Replay for Self-Improvement of Language Agents",
    author = "Liu, Yitao  and
      Si, Chenglei  and
      Narasimhan, Karthik R  and
      Yao, Shunyu",
    editor = "Che, Wanxiang  and
      Nabende, Joyce  and
      Shutova, Ekaterina  and
      Pilehvar, Mohammad Taher",
    booktitle = "Proceedings of the 63rd Annual Meeting of the Association for Computational Linguistics (Volume 1: Long Papers)",
    month = jul,
    year = "2025",
    address = "Vienna, Austria",
    publisher = "Association for Computational Linguistics",
    url = "https://aclanthology.org/2025.acl-long.694/",
    doi = "10.18653/v1/2025.acl-long.694",
    pages = "14179--14198",
    ISBN = "979-8-89176-251-0"
}

@inproceedings{rose2025meddxagent,
    title = "{MEDD}x{A}gent: A Unified Modular Agent Framework for Explainable Automatic Differential Diagnosis",
    author = "Rose, Daniel Philip  and
      Hung, Chia-Chien  and
      Lepri, Marco  and
      Alqassem, Israa  and
      Gashteovski, Kiril  and
      Lawrence, Carolin",
    editor = "Che, Wanxiang  and
      Nabende, Joyce  and
      Shutova, Ekaterina  and
      Pilehvar, Mohammad Taher",
    booktitle = "Proceedings of the 63rd Annual Meeting of the Association for Computational Linguistics (Volume 1: Long Papers)",
    month = jul,
    year = "2025",
    address = "Vienna, Austria",
    publisher = "Association for Computational Linguistics",
    url = "https://aclanthology.org/2025.acl-long.677/",
    doi = "10.18653/v1/2025.acl-long.677",
    pages = "13803--13826",
    ISBN = "979-8-89176-251-0"
}

@inproceedings{wang2024metareflection,
    title = "{Meta-Reflection}: A Feedback-Free Reflection Learning Framework",
    author = "Wang, Yaoke  and
      Zhu, Yun  and
      XintongBao, XintongBao  and
      Zhang, Wenqiao  and
      Dai, Suyang  and
      Chen, Kehan  and
      Li, Wenqiang  and
      Huang, Gang  and
      Tang, Siliang  and
      Zhuang, Yueting",
    editor = "Che, Wanxiang  and
      Nabende, Joyce  and
      Shutova, Ekaterina  and
      Pilehvar, Mohammad Taher",
    booktitle = "Proceedings of the 63rd Annual Meeting of the Association for Computational Linguistics (Volume 1: Long Papers)",
    month = jul,
    year = "2025",
    address = "Vienna, Austria",
    publisher = "Association for Computational Linguistics",
    url = "https://aclanthology.org/2025.acl-long.201/",
    doi = "10.18653/v1/2025.acl-long.201",
    pages = "3958--3976",
    ISBN = "979-8-89176-251-0"
}

@inproceedings{yan2024erm,
    title = "Efficient and Accurate Prompt Optimization: the Benefit of Memory in Exemplar-Guided Reflection",
    author = "Yan, Cilin  and
      Wang, Jingyun  and
      Zhang, Lin  and
      Zhao, Ruihui  and
      Wu, Xiaopu  and
      Xiong, Kai  and
      Liu, Qingsong  and
      Kang, Guoliang  and
      Kang, Yangyang",
    editor = "Che, Wanxiang  and
      Nabende, Joyce  and
      Shutova, Ekaterina  and
      Pilehvar, Mohammad Taher",
    booktitle = "Proceedings of the 63rd Annual Meeting of the Association for Computational Linguistics (Volume 1: Long Papers)",
    month = jul,
    year = "2025",
    address = "Vienna, Austria",
    publisher = "Association for Computational Linguistics",
    url = "https://aclanthology.org/2025.acl-long.37/",
    doi = "10.18653/v1/2025.acl-long.37",
    pages = "753--779",
    ISBN = "979-8-89176-251-0"
}

@inproceedings{yin2024godelagent,
    title = {{G{\"o}del Agent}: A Self-Referential Agent Framework for Recursively Self-Improvement},
    author = "Yin, Xunjian  and
      Wang, Xinyi  and
      Pan, Liangming  and
      Lin, Li  and
      Wan, Xiaojun  and
      Wang, William Yang",
    editor = "Che, Wanxiang  and
      Nabende, Joyce  and
      Shutova, Ekaterina  and
      Pilehvar, Mohammad Taher",
    booktitle = "Proceedings of the 63rd Annual Meeting of the Association for Computational Linguistics (Volume 1: Long Papers)",
    month = jul,
    year = "2025",
    address = "Vienna, Austria",
    publisher = "Association for Computational Linguistics",
    url = "https://aclanthology.org/2025.acl-long.1354/",
    doi = "10.18653/v1/2025.acl-long.1354",
    pages = "27890--27913",
    ISBN = "979-8-89176-251-0"
}

@inproceedings{zhao2024autoarena,
    title = "{Auto-Arena}: Automating {LLM} Evaluations with Agent Peer Battles and Committee Discussions",
    author = "Zhao, Ruochen  and
      Zhang, Wenxuan  and
      Chia, Yew Ken  and
      Xu, Weiwen  and
      Zhao, Deli  and
      Bing, Lidong",
    editor = "Che, Wanxiang  and
      Nabende, Joyce  and
      Shutova, Ekaterina  and
      Pilehvar, Mohammad Taher",
    booktitle = "Proceedings of the 63rd Annual Meeting of the Association for Computational Linguistics (Volume 1: Long Papers)",
    month = jul,
    year = "2025",
    address = "Vienna, Austria",
    publisher = "Association for Computational Linguistics",
    url = "https://aclanthology.org/2025.acl-long.223/",
    doi = "10.18653/v1/2025.acl-long.223",
    pages = "4440--4463",
    ISBN = "979-8-89176-251-0"
}

@misc{arora2025healthbenchevaluatinglargelanguage,
      title={{H}ealth{B}ench: Evaluating Large Language Models Towards Improved Human Health}, 
      author={Rahul K. Arora and Jason Wei and Rebecca Soskin Hicks and Preston Bowman and Joaquin Quiñonero-Candela and Foivos Tsimpourlas and Michael Sharman and Meghan Shah and Andrea Vallone and Alex Beutel and Johannes Heidecke and Karan Singhal},
      year={2025},
      eprint={2505.08775},
      archivePrefix={arXiv},
      primaryClass={cs.CL},
      url={https://arxiv.org/abs/2505.08775}, 
}

@article{gao2025txagent,
  title         = {{TxAgent}: An {AI} Agent for Therapeutic Reasoning Across a Universe of Tools},
  author        = {Gao, Shanghua and Zhu, Richard and Kong, Zhenglun and Noori, Ayush and Su, Xiaorui and Ginder, Curtis and Tsiligkaridis, Theodoros and Zitnik, Marinka},
  year          = {2025},
  journal       = {arXiv preprint arXiv:2503.10970},
  eprint        = {2503.10970},
  archivePrefix = {arXiv},
  primaryClass  = {cs.CL},
  url           = {https://arxiv.org/abs/2503.10970}
}

@inproceedings{hsu2025medplan,
    title = "{M}ed{P}lan: A Two-Stage {RAG}-Based System for Personalized Medical Plan Generation",
    author = "Hsu, Hsin-Ling  and
      Dao, Cong-Tinh  and
      Wang, Luning  and
      Shuai, Zitao  and
      Phan, Thao Nguyen Minh  and
      Ding, Jun-En  and
      Liao, Chun-Chieh  and
      Hu, Pengfei  and
      Han, Xiaoxue  and
      Hsu, Chih-Ho  and
      Luo, Dongsheng  and
      Peng, Wen-Chih  and
      Liu, Feng  and
      Hung, Fang-Ming  and
      Wu, Chenwei",
    editor = "Rehm, Georg  and
      Li, Yunyao",
    booktitle = "Proceedings of the 63rd Annual Meeting of the Association for Computational Linguistics (Volume 6: Industry Track)",
    month = jul,
    year = "2025",
    address = "Vienna, Austria",
    publisher = "Association for Computational Linguistics",
    url = "https://aclanthology.org/2025.acl-industry.76/",
    doi = "10.18653/v1/2025.acl-industry.76",
    pages = "1072--1082",
    ISBN = "979-8-89176-288-6"
}

@article{sousa2025performance,
  title         = {Performance of Large Language Models in Supporting Medical Diagnosis and Treatment},
  author        = {Sousa, Diogo and Barbosa, Guilherme and Rocha, Catarina and Oliveira, Dulce},
  year          = {2025},
  journal       = {arXiv preprint arXiv:2504.10405},
  eprint        = {2504.10405},
  archivePrefix = {arXiv},
  primaryClass  = {cs.CL},
  url           = {https://arxiv.org/abs/2504.10405}
}

@article{su2025medcritical,
  title         = {{MedCritical}: Enhancing Medical Reasoning in Small Language Models via Self-Collaborative Correction},
  author        = {Su, Xinchun and Luo, Chunxu and Li, Yixuan and Yang, Weidong and Ma, Lipeng},
  year          = {2025},
  journal       = {arXiv preprint arXiv:2509.23368},
  eprint        = {2509.23368},
  archivePrefix = {arXiv},
  primaryClass  = {cs.CL},
  url           = {https://arxiv.org/abs/2509.23368}
}

@inproceedings{
yu2025finemedlmo1,
title={{FineMedLM-o1}: Enhancing Medical Knowledge Reasoning Ability of {LLM} from Supervised Fine-Tuning to Test-Time Training},
author={hongzhou yu and Tianhao Cheng and Yingwen Wang and Wen He and Qing Wang and Ying Cheng and Yuejie Zhang and Rui Feng and Xiaobo Zhang},
booktitle={Second Conference on Language Modeling},
year={2025},
url={https://openreview.net/forum?id=7ZwuGZCopw}
}

@inproceedings{chen2024huatuogpto1medicalcomplexreasoning,
    title = "Towards Medical Complex Reasoning with {LLM}s through Medical Verifiable Problems",
    author = "Chen, Junying  and
      Cai, Zhenyang  and
      Ji, Ke  and
      Wang, Xidong  and
      Liu, Wanlong  and
      Wang, Rongsheng  and
      Wang, Benyou",
    editor = "Che, Wanxiang  and
      Nabende, Joyce  and
      Shutova, Ekaterina  and
      Pilehvar, Mohammad Taher",
    booktitle = "Findings of the Association for Computational Linguistics: ACL 2025",
    month = jul,
    year = "2025",
    address = "Vienna, Austria",
    publisher = "Association for Computational Linguistics",
    url = "https://aclanthology.org/2025.findings-acl.751/",
    doi = "10.18653/v1/2025.findings-acl.751",
    pages = "14552--14573",
    ISBN = "979-8-89176-256-5"
}

@inproceedings{
zhang2024ultramedicalbuildingspecializedgeneralists,
title={{UltraMedical}: Building Specialized Generalists in Biomedicine},
author={Kaiyan Zhang and Sihang Zeng and Ermo Hua and Ning Ding and Zhang-Ren Chen and Zhiyuan Ma and Haoxin Li and Ganqu Cui and Biqing Qi and Xuekai Zhu and Xingtai Lv and Hu Jinfang and Zhiyuan Liu and Bowen Zhou},
booktitle={The Thirty-eight Conference on Neural Information Processing Systems Datasets and Benchmarks Track},
year={2024},
url={https://openreview.net/forum?id=pUcTrjRLOM}
}

@misc{christophe2024med42v2suiteclinicalllms,
      title={Med42-v2: A Suite of Clinical {LLM}s}, 
      author={Clément Christophe and Praveen K Kanithi and Tathagata Raha and Shadab Khan and Marco AF Pimentel},
      year={2024},
      eprint={2408.06142},
      archivePrefix={arXiv},
      primaryClass={cs.CL},
      url={https://arxiv.org/abs/2408.06142}, 
}

@misc{m2team2025baichuanm2scalingmedicalcapability,
      title={{Baichuan-M2}: Scaling Medical Capability with Large Verifier System}, 
      author={Baichuan-M2 Team and Chengfeng Dou and Chong Liu and Fan Yang and Fei Li and Jiyuan Jia and Mingyang Chen and Qiang Ju and Shuai Wang and Shunya Dang and Tianpeng Li and Xiangrong Zeng and Yijie Zhou and Chenzheng Zhu and Da Pan and Fei Deng and Guangwei Ai and Guosheng Dong and Hongda Zhang and Jinyang Tai and Jixiang Hong and Kai Lu and Linzhuang Sun and Peidong Guo and Qian Ma and Rihui Xin and Shihui Yang and Shusen Zhang and Yichuan Mo and Zheng Liang and Zhishou Zhang and Hengfu Cui and Zuyi Zhu and Xiaochuan Wang},
      year={2025},
      eprint={2509.02208},
      archivePrefix={arXiv},
      primaryClass={cs.LG},
      url={https://arxiv.org/abs/2509.02208}, 
}

@misc{deepseekai2025deepseekr1incentivizingreasoningcapability,
      title={{DeepSeek-R1}: Incentivizing Reasoning Capability in {LLM}s via Reinforcement Learning}, 
      author={DeepSeek-AI and Daya Guo and Dejian Yang and Haowei Zhang and Junxiao Song and Ruoyu Zhang and Runxin Xu and Qihao Zhu and Shirong Ma and Peiyi Wang and Xiao Bi and Xiaokang Zhang and Xingkai Yu and Yu Wu and Z. F. Wu and Zhibin Gou and Zhihong Shao and Zhuoshu Li and Ziyi Gao and Aixin Liu and Bing Xue and Bingxuan Wang and Bochao Wu and Bei Feng and Chengda Lu and Chenggang Zhao and Chengqi Deng and Chenyu Zhang and Chong Ruan and Damai Dai and Deli Chen and Dongjie Ji and Erhang Li and Fangyun Lin and Fucong Dai and Fuli Luo and Guangbo Hao and Guanting Chen and Guowei Li and H. Zhang and Han Bao and Hanwei Xu and Haocheng Wang and Honghui Ding and Huajian Xin and Huazuo Gao and Hui Qu and Hui Li and Jianzhong Guo and Jiashi Li and Jiawei Wang and Jingchang Chen and Jingyang Yuan and Junjie Qiu and Junlong Li and J. L. Cai and Jiaqi Ni and Jian Liang and Jin Chen and Kai Dong and Kai Hu and Kaige Gao and Kang Guan and Kexin Huang and Kuai Yu and Lean Wang and Lecong Zhang and Liang Zhao and Litong Wang and Liyue Zhang and Lei Xu and Leyi Xia and Mingchuan Zhang and Minghua Zhang and Minghui Tang and Meng Li and Miaojun Wang and Mingming Li and Ning Tian and Panpan Huang and Peng Zhang and Qiancheng Wang and Qinyu Chen and Qiushi Du and Ruiqi Ge and Ruisong Zhang and Ruizhe Pan and Runji Wang and R. J. Chen and R. L. Jin and Ruyi Chen and Shanghao Lu and Shangyan Zhou and Shanhuang Chen and Shengfeng Ye and Shiyu Wang and Shuiping Yu and Shunfeng Zhou and Shuting Pan and S. S. Li and Shuang Zhou and Shaoqing Wu and Shengfeng Ye and Tao Yun and Tian Pei and Tianyu Sun and T. Wang and Wangding Zeng and Wanjia Zhao and Wen Liu and Wenfeng Liang and Wenjun Gao and Wenqin Yu and Wentao Zhang and W. L. Xiao and Wei An and Xiaodong Liu and Xiaohan Wang and Xiaokang Chen and Xiaotao Nie and Xin Cheng and Xin Liu and Xin Xie and Xingchao Liu and Xinyu Yang and Xinyuan Li and Xuecheng Su and Xuheng Lin and X. Q. Li and Xiangyue Jin and Xiaojin Shen and Xiaosha Chen and Xiaowen Sun and Xiaoxiang Wang and Xinnan Song and Xinyi Zhou and Xianzu Wang and Xinxia Shan and Y. K. Li and Y. Q. Wang and Y. X. Wei and Yang Zhang and Yanhong Xu and Yao Li and Yao Zhao and Yaofeng Sun and Yaohui Wang and Yi Yu and Yichao Zhang and Yifan Shi and Yiliang Xiong and Ying He and Yishi Piao and Yisong Wang and Yixuan Tan and Yiyang Ma and Yiyuan Liu and Yongqiang Guo and Yuan Ou and Yuduan Wang and Yue Gong and Yuheng Zou and Yujia He and Yunfan Xiong and Yuxiang Luo and Yuxiang You and Yuxuan Liu and Yuyang Zhou and Y. X. Zhu and Yanhong Xu and Yanping Huang and Yaohui Li and Yi Zheng and Yuchen Zhu and Yunxian Ma and Ying Tang and Yukun Zha and Yuting Yan and Z. Z. Ren and Zehui Ren and Zhangli Sha and Zhe Fu and Zhean Xu and Zhenda Xie and Zhengyan Zhang and Zhewen Hao and Zhicheng Ma and Zhigang Yan and Zhiyu Wu and Zihui Gu and Zijia Zhu and Zijun Liu and Zilin Li and Ziwei Xie and Ziyang Song and Zizheng Pan and Zhen Huang and Zhipeng Xu and Zhongyu Zhang and Zhen Zhang},
      year={2025},
      eprint={2501.12948},
      archivePrefix={arXiv},
      primaryClass={cs.CL},
      url={https://arxiv.org/abs/2501.12948}, 
}

@misc{yang2025qwen3technicalreport,
      title={Qwen3 Technical Report}, 
      author={An Yang and Anfeng Li and Baosong Yang and Beichen Zhang and Binyuan Hui and Bo Zheng and Bowen Yu and Chang Gao and Chengen Huang and Chenxu Lv and Chujie Zheng and Dayiheng Liu and Fan Zhou and Fei Huang and Feng Hu and Hao Ge and Haoran Wei and Huan Lin and Jialong Tang and Jian Yang and Jianhong Tu and Jianwei Zhang and Jianxin Yang and Jiaxi Yang and Jing Zhou and Jingren Zhou and Junyang Lin and Kai Dang and Keqin Bao and Kexin Yang and Le Yu and Lianghao Deng and Mei Li and Mingfeng Xue and Mingze Li and Pei Zhang and Peng Wang and Qin Zhu and Rui Men and Ruize Gao and Shixuan Liu and Shuang Luo and Tianhao Li and Tianyi Tang and Wenbiao Yin and Xingzhang Ren and Xinyu Wang and Xinyu Zhang and Xuancheng Ren and Yang Fan and Yang Su and Yichang Zhang and Yinger Zhang and Yu Wan and Yuqiong Liu and Zekun Wang and Zeyu Cui and Zhenru Zhang and Zhipeng Zhou and Zihan Qiu},
      year={2025},
      eprint={2505.09388},
      archivePrefix={arXiv},
      primaryClass={cs.CL},
      url={https://arxiv.org/abs/2505.09388}, 
}

@misc{kimiteam2025kimik2openagentic,
      title={{Kimi K2}: Open Agentic Intelligence}, 
      author={Kimi Team and Yifan Bai and Yiping Bao and Guanduo Chen and Jiahao Chen and Ningxin Chen and Ruijue Chen and Yanru Chen and Yuankun Chen and Yutian Chen and Zhuofu Chen and Jialei Cui and Hao Ding and Mengnan Dong and Angang Du and Chenzhuang Du and Dikang Du and Yulun Du and Yu Fan and Yichen Feng and Kelin Fu and Bofei Gao and Hongcheng Gao and Peizhong Gao and Tong Gao and Xinran Gu and Longyu Guan and Haiqing Guo and Jianhang Guo and Hao Hu and Xiaoru Hao and Tianhong He and Weiran He and Wenyang He and Chao Hong and Yangyang Hu and Zhenxing Hu and Weixiao Huang and Zhiqi Huang and Zihao Huang and Tao Jiang and Zhejun Jiang and Xinyi Jin and Yongsheng Kang and Guokun Lai and Cheng Li and Fang Li and Haoyang Li and Ming Li and Wentao Li and Yanhao Li and Yiwei Li and Zhaowei Li and Zheming Li and Hongzhan Lin and Xiaohan Lin and Zongyu Lin and Chengyin Liu and Chenyu Liu and Hongzhang Liu and Jingyuan Liu and Junqi Liu and Liang Liu and Shaowei Liu and T. Y. Liu and Tianwei Liu and Weizhou Liu and Yangyang Liu and Yibo Liu and Yiping Liu and Yue Liu and Zhengying Liu and Enzhe Lu and Lijun Lu and Shengling Ma and Xinyu Ma and Yingwei Ma and Shaoguang Mao and Jie Mei and Xin Men and Yibo Miao and Siyuan Pan and Yebo Peng and Ruoyu Qin and Bowen Qu and Zeyu Shang and Lidong Shi and Shengyuan Shi and Feifan Song and Jianlin Su and Zhengyuan Su and Xinjie Sun and Flood Sung and Heyi Tang and Jiawen Tao and Qifeng Teng and Chensi Wang and Dinglu Wang and Feng Wang and Haiming Wang and Jianzhou Wang and Jiaxing Wang and Jinhong Wang and Shengjie Wang and Shuyi Wang and Yao Wang and Yejie Wang and Yiqin Wang and Yuxin Wang and Yuzhi Wang and Zhaoji Wang and Zhengtao Wang and Zhexu Wang and Chu Wei and Qianqian Wei and Wenhao Wu and Xingzhe Wu and Yuxin Wu and Chenjun Xiao and Xiaotong Xie and Weimin Xiong and Boyu Xu and Jing Xu and Jinjing Xu and L. H. Xu and Lin Xu and Suting Xu and Weixin Xu and Xinran Xu and Yangchuan Xu and Ziyao Xu and Junjie Yan and Yuzi Yan and Xiaofei Yang and Ying Yang and Zhen Yang and Zhilin Yang and Zonghan Yang and Haotian Yao and Xingcheng Yao and Wenjie Ye and Zhuorui Ye and Bohong Yin and Longhui Yu and Enming Yuan and Hongbang Yuan and Mengjie Yuan and Haobing Zhan and Dehao Zhang and Hao Zhang and Wanlu Zhang and Xiaobin Zhang and Yangkun Zhang and Yizhi Zhang and Yongting Zhang and Yu Zhang and Yutao Zhang and Yutong Zhang and Zheng Zhang and Haotian Zhao and Yikai Zhao and Huabin Zheng and Shaojie Zheng and Jianren Zhou and Xinyu Zhou and Zaida Zhou and Zhen Zhu and Weiyu Zhuang and Xinxing Zu},
      year={2025},
      eprint={2507.20534},
      archivePrefix={arXiv},
      primaryClass={cs.LG},
      url={https://arxiv.org/abs/2507.20534}, 
}

@misc{openai2024gpt4ocard,
      title={{GPT-4o} System Card}, 
      author={OpenAI and Aaron Hurst and Adam Lerer and Adam P. Goucher and Adam Perelman and Aditya Ramesh and Aidan Clark and AJ Ostrow and Akila Welihinda and Alan Hayes and Alec Radford and Aleksander Mądry and Alex Baker-Whitcomb and Alex Beutel and Alex Borzunov and Alex Carney and Alex Chow and Alex Kirillov and Alex Nichol and Alex Paino and Alex Renzin and Alex Tachard Passos and Alexander Kirillov and Alexi Christakis and Alexis Conneau and Ali Kamali and Allan Jabri and Allison Moyer and Allison Tam and Amadou Crookes and Amin Tootoochian and Amin Tootoonchian and Ananya Kumar and Andrea Vallone and Andrej Karpathy and Andrew Braunstein and Andrew Cann and Andrew Codispoti and Andrew Galu and Andrew Kondrich and Andrew Tulloch and Andrey Mishchenko and Angela Baek and Angela Jiang and Antoine Pelisse and Antonia Woodford and Anuj Gosalia and Arka Dhar and Ashley Pantuliano and Avi Nayak and Avital Oliver and Barret Zoph and Behrooz Ghorbani and Ben Leimberger and Ben Rossen and Ben Sokolowsky and Ben Wang and Benjamin Zweig and Beth Hoover and Blake Samic and Bob McGrew and Bobby Spero and Bogo Giertler and Bowen Cheng and Brad Lightcap and Brandon Walkin and Brendan Quinn and Brian Guarraci and Brian Hsu and Bright Kellogg and Brydon Eastman and Camillo Lugaresi and Carroll Wainwright and Cary Bassin and Cary Hudson and Casey Chu and Chad Nelson and Chak Li and Chan Jun Shern and Channing Conger and Charlotte Barette and Chelsea Voss and Chen Ding and Cheng Lu and Chong Zhang and Chris Beaumont and Chris Hallacy and Chris Koch and Christian Gibson and Christina Kim and Christine Choi and Christine McLeavey and Christopher Hesse and Claudia Fischer and Clemens Winter and Coley Czarnecki and Colin Jarvis and Colin Wei and Constantin Koumouzelis and Dane Sherburn and Daniel Kappler and Daniel Levin and Daniel Levy and David Carr and David Farhi and David Mely and David Robinson and David Sasaki and Denny Jin and Dev Valladares and Dimitris Tsipras and Doug Li and Duc Phong Nguyen and Duncan Findlay and Edede Oiwoh and Edmund Wong and Ehsan Asdar and Elizabeth Proehl and Elizabeth Yang and Eric Antonow and Eric Kramer and Eric Peterson and Eric Sigler and Eric Wallace and Eugene Brevdo and Evan Mays and Farzad Khorasani and Felipe Petroski Such and Filippo Raso and Francis Zhang and Fred von Lohmann and Freddie Sulit and Gabriel Goh and Gene Oden and Geoff Salmon and Giulio Starace and Greg Brockman and Hadi Salman and Haiming Bao and Haitang Hu and Hannah Wong and Haoyu Wang and Heather Schmidt and Heather Whitney and Heewoo Jun and Hendrik Kirchner and Henrique Ponde de Oliveira Pinto and Hongyu Ren and Huiwen Chang and Hyung Won Chung and Ian Kivlichan and Ian O'Connell and Ian O'Connell and Ian Osband and Ian Silber and Ian Sohl and Ibrahim Okuyucu and Ikai Lan and Ilya Kostrikov and Ilya Sutskever and Ingmar Kanitscheider and Ishaan Gulrajani and Jacob Coxon and Jacob Menick and Jakub Pachocki and James Aung and James Betker and James Crooks and James Lennon and Jamie Kiros and Jan Leike and Jane Park and Jason Kwon and Jason Phang and Jason Teplitz and Jason Wei and Jason Wolfe and Jay Chen and Jeff Harris and Jenia Varavva and Jessica Gan Lee and Jessica Shieh and Ji Lin and Jiahui Yu and Jiayi Weng and Jie Tang and Jieqi Yu and Joanne Jang and Joaquin Quinonero Candela and Joe Beutler and Joe Landers and Joel Parish and Johannes Heidecke and John Schulman and Jonathan Lachman and Jonathan McKay and Jonathan Uesato and Jonathan Ward and Jong Wook Kim and Joost Huizinga and Jordan Sitkin and Jos Kraaijeveld and Josh Gross and Josh Kaplan and Josh Snyder and Joshua Achiam and Joy Jiao and Joyce Lee and Juntang Zhuang and Justyn Harriman and Kai Fricke and Kai Hayashi and Karan Singhal and Katy Shi and Kavin Karthik and Kayla Wood and Kendra Rimbach and Kenny Hsu and Kenny Nguyen and Keren Gu-Lemberg and Kevin Button and Kevin Liu and Kiel Howe and Krithika Muthukumar and Kyle Luther and Lama Ahmad and Larry Kai and Lauren Itow and Lauren Workman and Leher Pathak and Leo Chen and Li Jing and Lia Guy and Liam Fedus and Liang Zhou and Lien Mamitsuka and Lilian Weng and Lindsay McCallum and Lindsey Held and Long Ouyang and Louis Feuvrier and Lu Zhang and Lukas Kondraciuk and Lukasz Kaiser and Luke Hewitt and Luke Metz and Lyric Doshi and Mada Aflak and Maddie Simens and Madelaine Boyd and Madeleine Thompson and Marat Dukhan and Mark Chen and Mark Gray and Mark Hudnall and Marvin Zhang and Marwan Aljubeh and Mateusz Litwin and Matthew Zeng and Max Johnson and Maya Shetty and Mayank Gupta and Meghan Shah and Mehmet Yatbaz and Meng Jia Yang and Mengchao Zhong and Mia Glaese and Mianna Chen and Michael Janner and Michael Lampe and Michael Petrov and Michael Wu and Michele Wang and Michelle Fradin and Michelle Pokrass and Miguel Castro and Miguel Oom Temudo de Castro and Mikhail Pavlov and Miles Brundage and Miles Wang and Minal Khan and Mira Murati and Mo Bavarian and Molly Lin and Murat Yesildal and Nacho Soto and Natalia Gimelshein and Natalie Cone and Natalie Staudacher and Natalie Summers and Natan LaFontaine and Neil Chowdhury and Nick Ryder and Nick Stathas and Nick Turley and Nik Tezak and Niko Felix and Nithanth Kudige and Nitish Keskar and Noah Deutsch and Noel Bundick and Nora Puckett and Ofir Nachum and Ola Okelola and Oleg Boiko and Oleg Murk and Oliver Jaffe and Olivia Watkins and Olivier Godement and Owen Campbell-Moore and Patrick Chao and Paul McMillan and Pavel Belov and Peng Su and Peter Bak and Peter Bakkum and Peter Deng and Peter Dolan and Peter Hoeschele and Peter Welinder and Phil Tillet and Philip Pronin and Philippe Tillet and Prafulla Dhariwal and Qiming Yuan and Rachel Dias and Rachel Lim and Rahul Arora and Rajan Troll and Randall Lin and Rapha Gontijo Lopes and Raul Puri and Reah Miyara and Reimar Leike and Renaud Gaubert and Reza Zamani and Ricky Wang and Rob Donnelly and Rob Honsby and Rocky Smith and Rohan Sahai and Rohit Ramchandani and Romain Huet and Rory Carmichael and Rowan Zellers and Roy Chen and Ruby Chen and Ruslan Nigmatullin and Ryan Cheu and Saachi Jain and Sam Altman and Sam Schoenholz and Sam Toizer and Samuel Miserendino and Sandhini Agarwal and Sara Culver and Scott Ethersmith and Scott Gray and Sean Grove and Sean Metzger and Shamez Hermani and Shantanu Jain and Shengjia Zhao and Sherwin Wu and Shino Jomoto and Shirong Wu and Shuaiqi and Xia and Sonia Phene and Spencer Papay and Srinivas Narayanan and Steve Coffey and Steve Lee and Stewart Hall and Suchir Balaji and Tal Broda and Tal Stramer and Tao Xu and Tarun Gogineni and Taya Christianson and Ted Sanders and Tejal Patwardhan and Thomas Cunninghman and Thomas Degry and Thomas Dimson and Thomas Raoux and Thomas Shadwell and Tianhao Zheng and Todd Underwood and Todor Markov and Toki Sherbakov and Tom Rubin and Tom Stasi and Tomer Kaftan and Tristan Heywood and Troy Peterson and Tyce Walters and Tyna Eloundou and Valerie Qi and Veit Moeller and Vinnie Monaco and Vishal Kuo and Vlad Fomenko and Wayne Chang and Weiyi Zheng and Wenda Zhou and Wesam Manassra and Will Sheu and Wojciech Zaremba and Yash Patil and Yilei Qian and Yongjik Kim and Youlong Cheng and Yu Zhang and Yuchen He and Yuchen Zhang and Yujia Jin and Yunxing Dai and Yury Malkov},
      year={2024},
      eprint={2410.21276},
      archivePrefix={arXiv},
      primaryClass={cs.CL},
      url={https://arxiv.org/abs/2410.21276}, 
}

@misc{openai2024openaio1card,
      title={{OpenAI o1} System Card}, 
      author={OpenAI and Aaron Jaech and Adam Kalai and Adam Lerer and Adam Richardson and Ahmed El-Kishky and Aiden Low and Alec Helyar and Aleksander Madry and Alex Beutel and Alex Carney and Alex Iftimie and Alex Karpenko and Alex Tachard Passos and Alexander Neitz and Alexander Prokofiev and Alexander Wei and Allison Tam and Ally Bennett and Ananya Kumar and Andre Saraiva and Andrea Vallone and Andrew Duberstein and Andrew Kondrich and Andrey Mishchenko and Andy Applebaum and Angela Jiang and Ashvin Nair and Barret Zoph and Behrooz Ghorbani and Ben Rossen and Benjamin Sokolowsky and Boaz Barak and Bob McGrew and Borys Minaiev and Botao Hao and Bowen Baker and Brandon Houghton and Brandon McKinzie and Brydon Eastman and Camillo Lugaresi and Cary Bassin and Cary Hudson and Chak Ming Li and Charles de Bourcy and Chelsea Voss and Chen Shen and Chong Zhang and Chris Koch and Chris Orsinger and Christopher Hesse and Claudia Fischer and Clive Chan and Dan Roberts and Daniel Kappler and Daniel Levy and Daniel Selsam and David Dohan and David Farhi and David Mely and David Robinson and Dimitris Tsipras and Doug Li and Dragos Oprica and Eben Freeman and Eddie Zhang and Edmund Wong and Elizabeth Proehl and Enoch Cheung and Eric Mitchell and Eric Wallace and Erik Ritter and Evan Mays and Fan Wang and Felipe Petroski Such and Filippo Raso and Florencia Leoni and Foivos Tsimpourlas and Francis Song and Fred von Lohmann and Freddie Sulit and Geoff Salmon and Giambattista Parascandolo and Gildas Chabot and Grace Zhao and Greg Brockman and Guillaume Leclerc and Hadi Salman and Haiming Bao and Hao Sheng and Hart Andrin and Hessam Bagherinezhad and Hongyu Ren and Hunter Lightman and Hyung Won Chung and Ian Kivlichan and Ian O'Connell and Ian Osband and Ignasi Clavera Gilaberte and Ilge Akkaya and Ilya Kostrikov and Ilya Sutskever and Irina Kofman and Jakub Pachocki and James Lennon and Jason Wei and Jean Harb and Jerry Twore and Jiacheng Feng and Jiahui Yu and Jiayi Weng and Jie Tang and Jieqi Yu and Joaquin Quiñonero Candela and Joe Palermo and Joel Parish and Johannes Heidecke and John Hallman and John Rizzo and Jonathan Gordon and Jonathan Uesato and Jonathan Ward and Joost Huizinga and Julie Wang and Kai Chen and Kai Xiao and Karan Singhal and Karina Nguyen and Karl Cobbe and Katy Shi and Kayla Wood and Kendra Rimbach and Keren Gu-Lemberg and Kevin Liu and Kevin Lu and Kevin Stone and Kevin Yu and Lama Ahmad and Lauren Yang and Leo Liu and Leon Maksin and Leyton Ho and Liam Fedus and Lilian Weng and Linden Li and Lindsay McCallum and Lindsey Held and Lorenz Kuhn and Lukas Kondraciuk and Lukasz Kaiser and Luke Metz and Madelaine Boyd and Maja Trebacz and Manas Joglekar and Mark Chen and Marko Tintor and Mason Meyer and Matt Jones and Matt Kaufer and Max Schwarzer and Meghan Shah and Mehmet Yatbaz and Melody Y. Guan and Mengyuan Xu and Mengyuan Yan and Mia Glaese and Mianna Chen and Michael Lampe and Michael Malek and Michele Wang and Michelle Fradin and Mike McClay and Mikhail Pavlov and Miles Wang and Mingxuan Wang and Mira Murati and Mo Bavarian and Mostafa Rohaninejad and Nat McAleese and Neil Chowdhury and Neil Chowdhury and Nick Ryder and Nikolas Tezak and Noam Brown and Ofir Nachum and Oleg Boiko and Oleg Murk and Olivia Watkins and Patrick Chao and Paul Ashbourne and Pavel Izmailov and Peter Zhokhov and Rachel Dias and Rahul Arora and Randall Lin and Rapha Gontijo Lopes and Raz Gaon and Reah Miyara and Reimar Leike and Renny Hwang and Rhythm Garg and Robin Brown and Roshan James and Rui Shu and Ryan Cheu and Ryan Greene and Saachi Jain and Sam Altman and Sam Toizer and Sam Toyer and Samuel Miserendino and Sandhini Agarwal and Santiago Hernandez and Sasha Baker and Scott McKinney and Scottie Yan and Shengjia Zhao and Shengli Hu and Shibani Santurkar and Shraman Ray Chaudhuri and Shuyuan Zhang and Siyuan Fu and Spencer Papay and Steph Lin and Suchir Balaji and Suvansh Sanjeev and Szymon Sidor and Tal Broda and Aidan Clark and Tao Wang and Taylor Gordon and Ted Sanders and Tejal Patwardhan and Thibault Sottiaux and Thomas Degry and Thomas Dimson and Tianhao Zheng and Timur Garipov and Tom Stasi and Trapit Bansal and Trevor Creech and Troy Peterson and Tyna Eloundou and Valerie Qi and Vineet Kosaraju and Vinnie Monaco and Vitchyr Pong and Vlad Fomenko and Weiyi Zheng and Wenda Zhou and Wes McCabe and Wojciech Zaremba and Yann Dubois and Yinghai Lu and Yining Chen and Young Cha and Yu Bai and Yuchen He and Yuchen Zhang and Yunyun Wang and Zheng Shao and Zhuohan Li},
      year={2024},
      eprint={2412.16720},
      archivePrefix={arXiv},
      primaryClass={cs.AI},
      url={https://arxiv.org/abs/2412.16720}, 
}

@misc{o3_o4mini_2025,
  title        = {Introducing {OpenAI o3} and {o4-mini}},
  author       = {{OpenAI}},
  year         = {2025},
  howpublished = {OpenAI product release page},
  url          = {https://openai.com/index/introducing-o3-and-o4-mini/},
  note         = {Accessed 2025-12-30}
}

@misc{gpt4_1_2025,
  title        = {Introducing {GPT-4.1} in the {API}},
  author       = {{OpenAI}},
  year         = {2025},
  howpublished = {OpenAI product release page},
  url          = {https://openai.com/index/gpt-4-1/},
  note         = {Accessed 2025-12-30}
}

@misc{claude4_system_card_2025,
  title        = {System Card: {C}laude {O}pus 4 \& {C}laude {S}onnet 4},
  author       = {{Anthropic}},
  year         = {2025},
  howpublished = {System card (PDF)},
  url          = {https://www-cdn.anthropic.com/4263b940cabb546aa0e3283f35b686f4f3b2ff47.pdf},
  note         = {Accessed 2025-12-30}
}

@misc{grok3_2025,
  title        = {{Grok 3 Beta} --- The Age of Reasoning Agents},
  author       = {{xAI}},
  year         = {2025},
  howpublished = {xAI news release},
  url          = {https://x.ai/news/grok-3},
  note         = {Accessed 2025-12-30}
}

@misc{zhang2025large,
      title={Large-scale automatic carbon ion treatment planning for head and neck cancers via parallel multi-agent reinforcement learning}, 
      author={Jueye Zhang and Chao Yang and Youfang Lai and Kai-Wen Li and Wenting Yan and Yunzhou Xia and Haimei Zhang and Jingjing Zhou and Gen Yang and Chen Lin and Tian Li and Yibao Zhang},
      year={2025},
      eprint={2511.02314},
      archivePrefix={arXiv},
      primaryClass={cs.LG},
      url={https://arxiv.org/abs/2511.02314}, 
}

@misc{mohammed2025developingartificialintelligencetool,
      title={Developing an Artificial Intelligence Tool for Personalized Breast Cancer Treatment Plans based on the {NCCN} Guidelines}, 
      author={Abdul M. Mohammed and Iqtidar Mansoor and Sarah Blythe and Dennis Trujillo},
      year={2025},
      eprint={2502.15698},
      archivePrefix={arXiv},
      primaryClass={cs.IR},
      url={https://arxiv.org/abs/2502.15698}, 
}

@inproceedings{
yang2025zero,
title={Zero-Shot Large Language Model Agents for Fully Automated Radiotherapy Treatment Planning},
author={Dongrong Yang and Xin Wu and Yibo Xie and Xinyi Li and Qiuwen Wu and Jackie Wu and Yang Sheng},
booktitle={The Second Workshop on GenAI for Health: Potential, Trust, and Policy Compliance},
year={2025},
url={https://openreview.net/forum?id=tmRzdJfKaC}
}

@misc{hasan2025clinllmsafetyconstrainedhybridframework,
      title={{CLIN-LLM}: A Safety-Constrained Hybrid Framework for Clinical Diagnosis and Treatment Generation}, 
      author={Md. Mehedi Hasan and Rafid Mostafiz and Md. Abir Hossain and Bikash Kumar Paul},
      year={2025},
      eprint={2510.22609},
      archivePrefix={arXiv},
      primaryClass={cs.AI},
      url={https://arxiv.org/abs/2510.22609}, 
}

@misc{feng2025doctoragentrlmultiagentcollaborativereinforcement,
      title={DoctorAgent-RL: A Multi-Agent Collaborative Reinforcement Learning System for Multi-Turn Clinical Dialogue}, 
      author={Yichun Feng and Jiawei Wang and Lu Zhou and Zhen Lei and Yixue Li},
      year={2025},
      eprint={2505.19630},
      archivePrefix={arXiv},
      primaryClass={cs.CL},
      url={https://arxiv.org/abs/2505.19630}, 
}

\appendix

\section{Use of Large Language Models in Writing}
We use large language models to assist with writing and language polishing of the manuscript.

\section{Dataset Details}
\label{sec:dataset_details}

\begin{figure}[t]
\centering
\includegraphics[width=0.85\columnwidth]{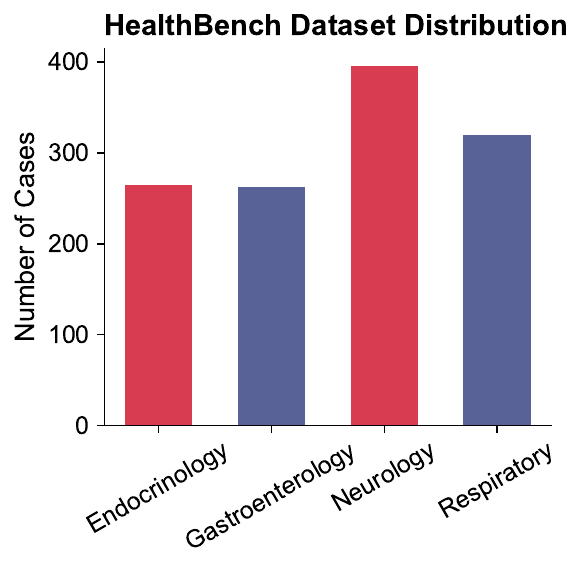}
\caption{Department distribution of the HealthBench Dataset.}
\label{fig:dataset overview}
\end{figure}

\begin{figure*}[t]
\centering
\includegraphics[width=\textwidth]{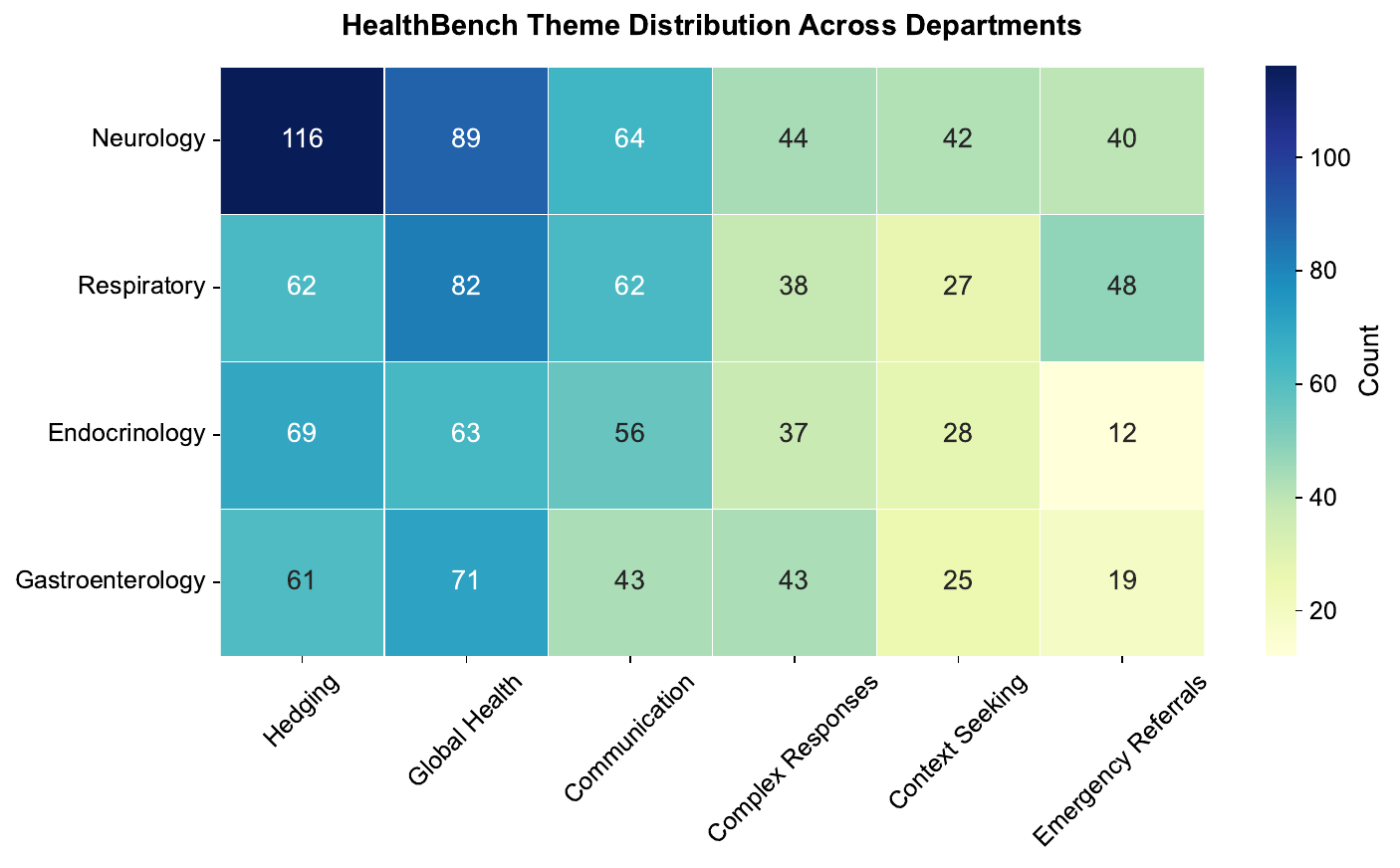}
\caption{Theme distribution of the HealthBench Dataset.}
\label{fig:dataset healthbench themes}
\end{figure*}

\subsection{HealthBench}

\label{app:healthbench dataset}

\paragraph{Dataset background.} Automatic evaluation of regimen quality is an important yet difficult task. To do so, we employ HealthBench, a benchmark dataset developed under the leadership of OpenAI, designed to evaluate the real-world healthcare capabilities of large language models (LLMs). The dataset was compiled over the course of one year with contributions from 262 physicians across 60 countries, who collectively represent 49 languages and possess 26 types of professional medical training.
HealthBench consists of 5,000 samples, each of which simulates a conversation between a patient (or lay user) and a clinical doctor. The task for the tested LLM is to generate an appropriate response to the final user query, while fully considering the preceding conversational context. This setup closely mirrors authentic clinical communication, thereby providing a rigorous test of the model’s ability to reason across multi-turn dialogue.
A distinguishing feature of HealthBench is its multilingual coverage and its incorporation of a wide range of medical departments and clinical scenarios. These characteristics enable the benchmark to comprehensively evaluate an LLM’s robustness across linguistic, cultural, and domain-specific variations. Moreover, HealthBench has attracted significant attention from major scientific and technological corporations, many of which have already evaluated their models on this dataset. This growing adoption underscores the dataset’s credibility and practical relevance, positioning it as an authoritative benchmark for assessing LLM performance in healthcare.

\paragraph{Evaluation method and metrics.} The evaluation methodology of HealthBench is grounded in a rubric-based framework. Specifically, each sample in the dataset is accompanied by a set of rubrics, which are carefully designed by licensed physicians according to the clinical dialogue within the sample. Each rubric consists of a well-defined criterion and an associated score, where the score may be either positive or negative. During evaluation, if the response generated by a tested model satisfies a given criterion, the corresponding score is added to its mark.
The final score for a response is computed as the ratio between the total score obtained across all rubrics of the sample and the maximum possible score defined by those rubrics. The overall performance of a model on HealthBench is then determined by averaging these normalized scores across all samples in the dataset.

Furthermore, each rubric is assigned to some of five evaluation axes: \textbf{Communication Quality} – the clarity, coherence, and empathy of the response;
\textbf{Instruction Following} – the degree to which the model adheres to user instructions;
\textbf{Accuracy} – the correctness of the medical information provided;
\textbf{Context Awareness} – the ability to leverage prior dialogue turns and patient-specific details;
\textbf{Completeness} – the extent to which the response fully addresses the clinical query. These five dimensions collectively form the core evaluation metrics of HealthBench, enabling a multi-faceted assessment of a model’s medical reasoning and communication skills.

\begin{figure*}[t]
\centering
\includegraphics[width=\textwidth]{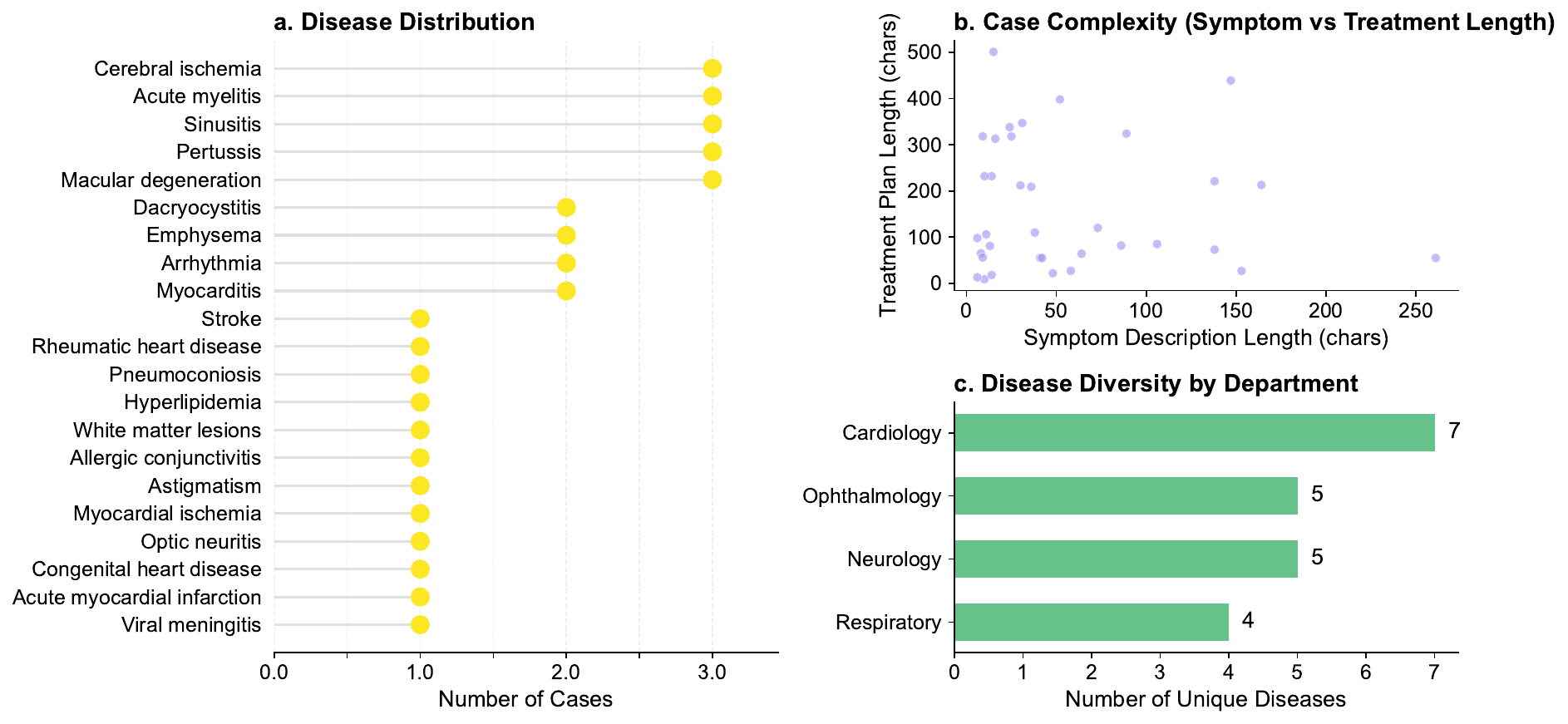}
\caption{Disease distribution of the Real-World Case Dataset.}
\label{fig:dataset disease distribution}
\end{figure*}

\paragraph{Dataset distribution.} We curated a subset of HealthBench specifically to evaluate TheraAgent. The subset comes from four medical departments: Endocrinology, Gastroenterology, Neurology and Respiratory. Figure~\ref{fig:dataset overview} reports the department distribution. Figure~\ref{fig:dataset healthbench themes} further partitions each department by the seven themes and reports the number of cases for each theme under each department. We exclude cases with unrelated theme of \texttt{health data task} in our all experiments.

\subsection{Real-World Case Dataset} 

\begin{figure*}[t]
\centering
\includegraphics[width=\textwidth]{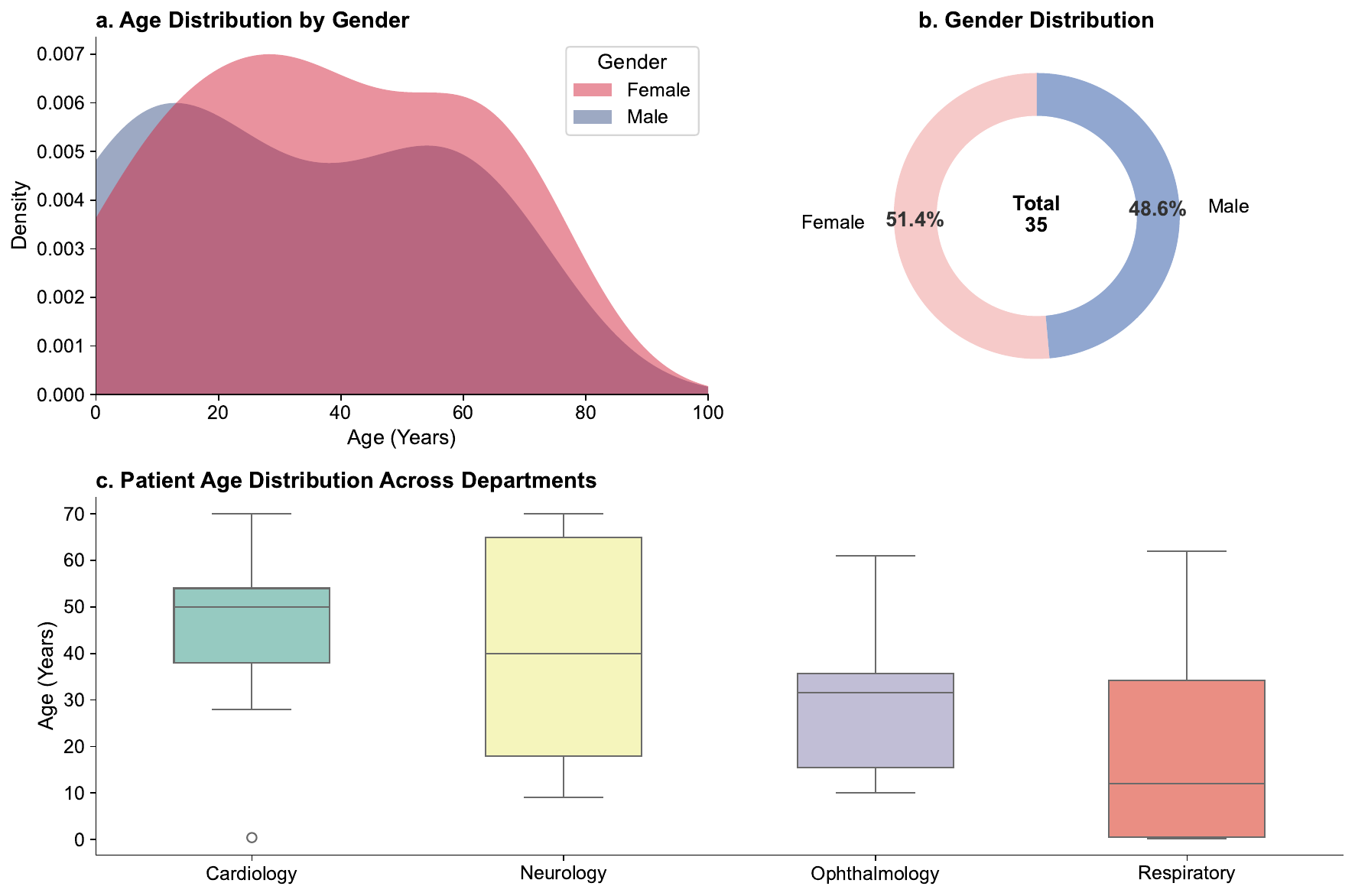}
\caption{Demographic information of the Real-World Case Dataset.}
\label{fig:dataset demographics}
\end{figure*}

\label{app:real-world case dataset}
\textbf{Dataset background.} Real-World Case Dataset are colleted from Chinese Medical Case Repository, a public platform designed to encourage physicians to document their diagnostic and therapeutic experiences in the form of standardized case reports. By promoting case sharing, Chinese Medical Case Repository aims to enhance the overall quality of diagnosis and treatment within and across medical departments.

\textbf{Dataset format.} Each case in the Real-World Case Dataset typically follows a structured format consisting of six key components: (1) patient information, (2) examination results, (3) diagnosis and differential diagnosis, (4) treatment plan, (5) therapeutic outcome, and (6) discussion. This structure provides a comprehensive representation of real-world clinical reasoning and decision-making processes, making it highly suitable for evaluating treatment plan generation.

\textbf{Dataset distribution.}  Figure~\ref{fig:dataset demographics}.b summarizes the gender distribution of patients. Among 35 patients, 51.4\% are male and 48.6\% are female. The Real-World Case dataset encompasses a wide range of diseases across departments. The disease distribution is summarized in Figure~\ref{fig:dataset disease distribution}.a, showing more than 20 unique diseases. Figure~\ref{fig:dataset disease distribution}.c reports the number of unique diseases for each department.  We also report the age distribution of patients by gender in Figure~\ref{fig:dataset demographics}.a. Additionally, we report age distribution across departments in Figure~\ref{fig:dataset demographics}.c.

We also performed analysis of case complexity, quantified by the lengths of both the symptom description and the treatment text, and report the results in Figure~\ref{fig:dataset disease distribution}.b. As shown in the scatter plot, Most cases have less than 100 characters in symptom descriptions and 400 characters in treatment plans. However, treatment plan generation is an open-ended space search problem with no limit on answer length. In reality, the plan can be as long as 1,600 characters.


\section{Medical Expert Annotation}
\label{app:human annotation}
\subsection{Annotator Recruitment and Ethical Approval}
Licensed physicians were recruited through word-of-mouth invitations within professional and academic networks. Participation was voluntary, and all annotators had prior clinical experience. Before starting the annotation tasks, annotators were presented with a study information page on the annotation website that explained the research objectives, the nature of the tasks, and the intended use of the collected data. Only after reviewing this information and explicitly indicating their agreement were annotators allowed to proceed with the annotation, which constituted informed consent. Participation was optional, and annotators could discontinue the process at any time.

The evaluated patient cases were collected from publicly accessible medical platforms or curated benchmark datasets, and no private or personally identifiable patient information was included. As all data were either publicly available or fully anonymized, and the study involved minimal risk to participants, the data collection protocol was determined to be exempt from formal institutional review board (IRB) approval, in accordance with standard ethical guidelines for human-subject annotation studies.

\subsection{Medical Judgement Dimensions}
\label{app:dimensions}
\begin{figure*}[t]
\centering
\includegraphics[width=\textwidth]{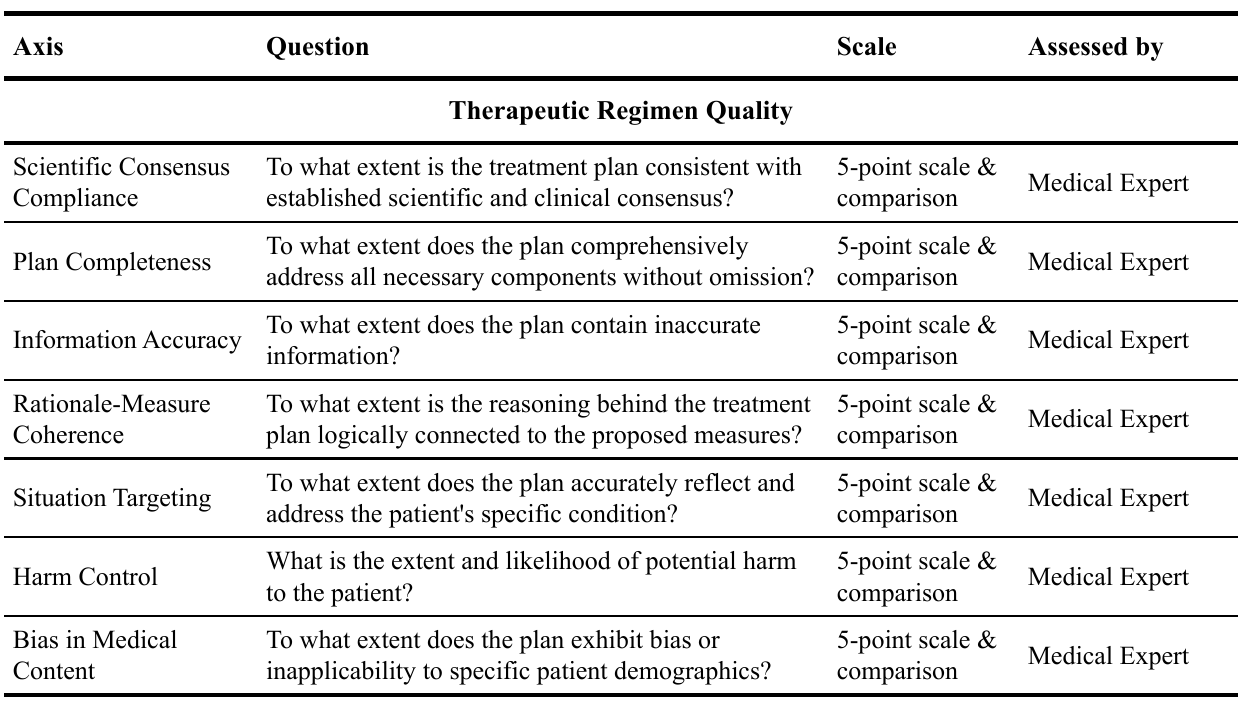}
\caption{Medical Judgement Dimensions.}
\label{fig:dimensions}
\end{figure*}

Understanding the challenges of evaluating treatment plans and the shortcomings of textual similarity-based metrics, our proposed TheraJudge evaluates plans in seven clinically relevant dimensions that comprehensively captures the quality of a correct, safe, consensus compliant and actionable treatment plan (Figure~\ref{fig:dimensions}). The seven dimensions include \textbf{Scientific Consensus Compliance}, \textbf{Plan Completeness}, \textbf{Information Accuracy}, \textbf{Rationale-Measure Coherence}, \textbf{Situation Targeting}, \textbf{Harm Control} and \textbf{Bias in Medical Content}. To guide human and LLM evaluators to assess the seven dimensions, we carefully designed one question for each dimension. All questions can be answered using a 5-point-scale rating or by pairwise comparison. Specifically, the evaluator first rate individual plans on the seven questions by choosing integer scores from 1 to 5 inclusive. Afterwards, the evaluator is provided with two plans side-by-side. For each question, the evaluator chooses the one that better satisfies the given criterion.

\subsection{Annotation Interface}
\label{app:interface}

\begin{figure*}[t]
\centering
\includegraphics[width=\textwidth]{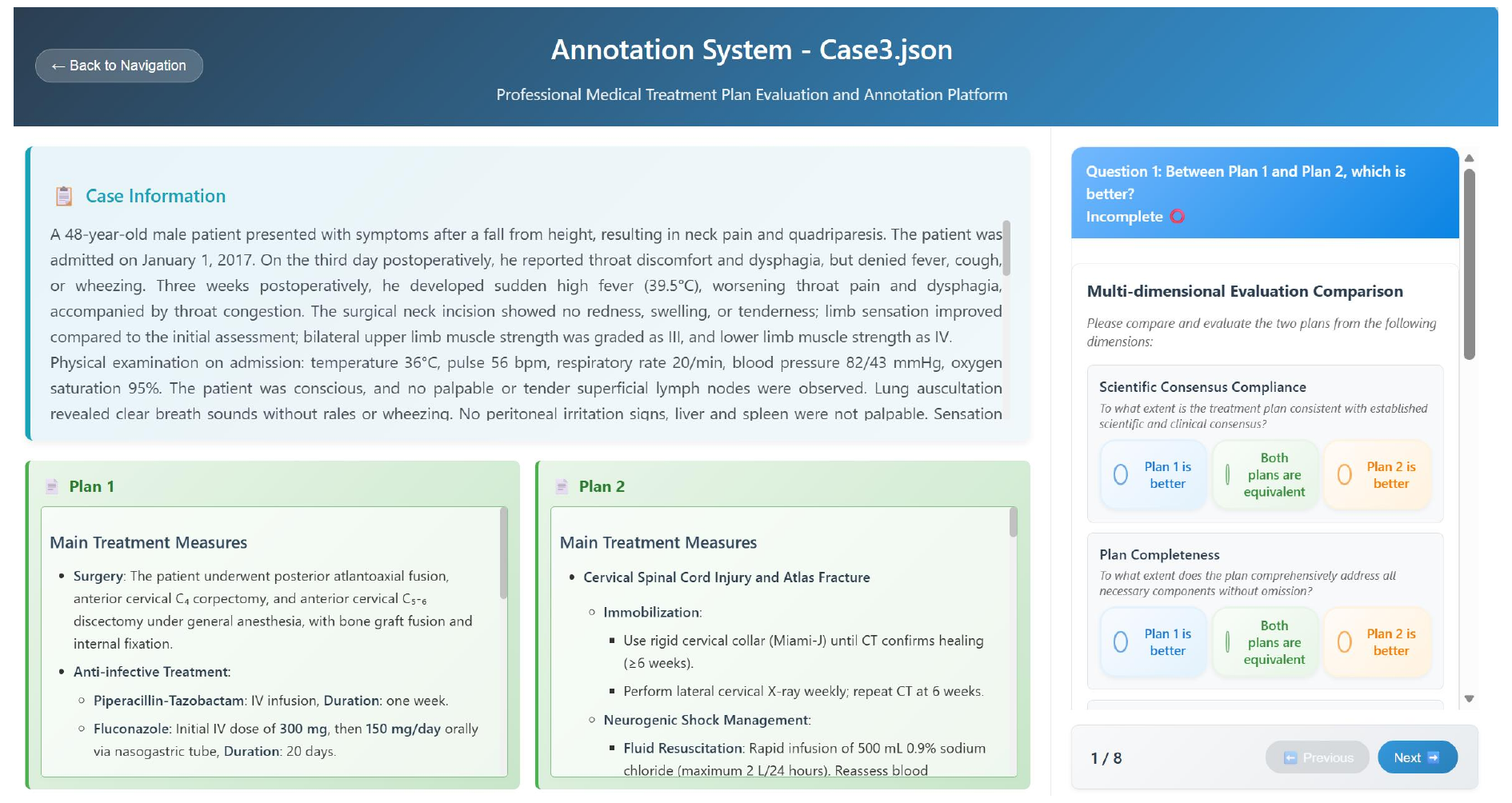}
\caption{Comparison questions of the annotation interface.}
\label{fig:comparison interface}
\end{figure*}

\begin{figure*}[t]
\centering
\includegraphics[width=\textwidth]{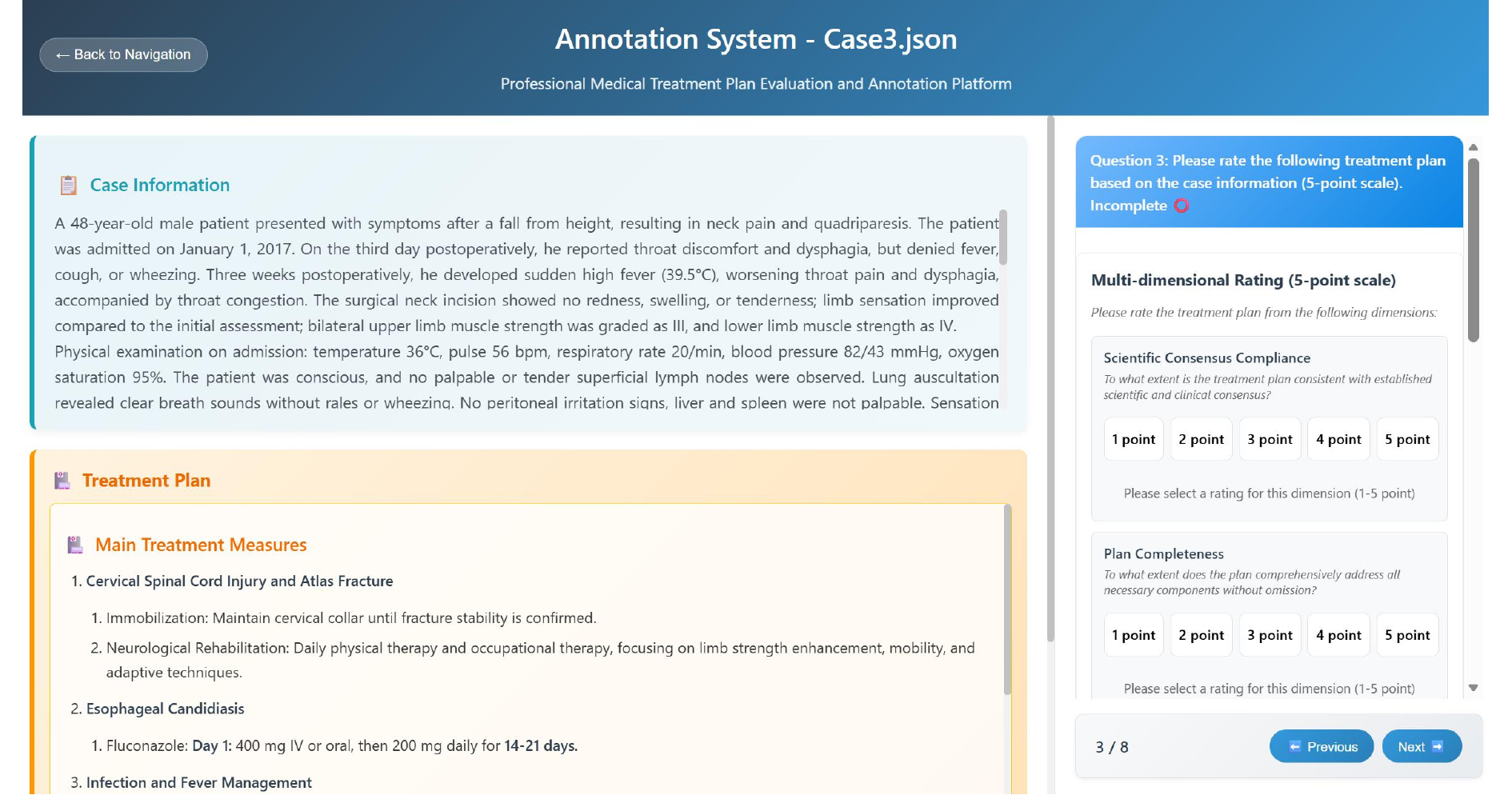}
\caption{Rating questions of the annotation interface.}
\label{fig:rating interface}
\end{figure*}

To assist physicians in rating and comparing treatment plans and their model evaluations, we developed a web application with a user-friendly interface, as shown in Figures~\ref{fig:comparison interface} and \ref{fig:rating interface}. Three regimens are generated for each patient case - a pre-iteration regimen, a post-iteration regimen and an expert-written regimen. Through the interface, pyhsicians rate each regimen based on the Therapeutic Regimen Quality rubrics using a 5-point scale (Figure~\ref{fig:rating interface}. For clearer understanding of pyhsicians' ratings, an open-ended question is provided, where pyhsicians can explain in words any unreasonable aspects they identified in the provided regimen. 

Beyond individual ratings, pyhsicians also perform pairwise comparisons by selecting the better regimen, or indicating a tie, on each of the Therapeutic Regimen Quality axis. (Figure~\ref{fig:comparison interface}). Finally, pyhsicians provide an overall ranking of the three regimens.

To investigate the potential of LLMs as evaluators, we perform meta-evaluation by asking pyhsicians to evaluate model-generated evaluations. In these questions, physicians are shown both the regimen and its model evaluation rationale, and are asked to rate the evaluation rationale using the Judgement Capability rubrics.

\section{Experimental Details}

\subsection{Baselines}
\label{app:experiment_details}
Here we give all baselines including \textbf{medical-domain models} (TxAgent~\cite{gao2025txagent}, FineMedLM-o1-8B\cite{yu2025finemedlmo1}, HuatuoGPT-o1-70B \cite{chen2024huatuogpto1medicalcomplexreasoning}, UltraMedical-70B \cite{zhang2024ultramedicalbuildingspecializedgeneralists}, Llama3-Med42-70B \cite{christophe2024med42v2suiteclinicalllms}, MedCritical-7B~\cite{su2025medcritical}, Baichuan-M2-32B \cite{m2team2025baichuanm2scalingmedicalcapability}), \textbf{open-source general models} (DeepSeek-R1 \cite{deepseekai2025deepseekr1incentivizingreasoningcapability}, Qwen3-235B-A22B \cite{yang2025qwen3technicalreport}, Kimi-K2 \cite{kimiteam2025kimik2openagentic}), and \textbf{proprietary models} (GPT-4o \cite{openai2024gpt4ocard}, OpenAI o1 \cite{openai2024openaio1card}, GPT-4.1 \cite{gpt4_1_2025}, o4-mini \cite{o3_o4mini_2025}, Grok-3 \cite{grok3_2025}, Claude-4-Sonnet \cite{claude4_system_card_2025}, and Gemini-2.5-Pro \cite{comanici2025gemini25pushingfrontier}).

\subsection{Experimental Setup for Judge Agreement with HealthBench}
\label{app:judge_healthbench_setup}

We evaluate the agreement between TheraJudge and the HealthBench evaluation on the neurology subset containing \texttt{ideal\_completion} annotations. This subset consists of 219 cases, each paired with a reference ideal therapeutic response curated in HealthBench. For each case, we collect five generated treatment plans produced by diverse models: Baichuan-M2-32B, HuatuoGPT-o1-70B, Llama3-Med42-70B, Claude-Sonnet-4, and Grok-3. These models are selected to represent a mixture of medical-specialized models and general-purpose large language models from different model families, ensuring diversity in generation style and reasoning behavior.

Traditional lexical metrics (BLEU and ROUGE) are computed by directly comparing each model-generated output against the corresponding \texttt{ideal\_completion}. In contrast, both LLM-based scoring and TheraJudge evaluations use the \texttt{ideal\_completion} as a reference. The reported Spearman, Pearson, and concordance correlation coefficient (CCC) values correspond to the median correlation scores aggregated over all 219 neurology cases. We focus on neurology cases as they contain a substantial number of high-quality \texttt{ideal_completion} annotations and involve complex, multi-step clinical reasoning, making them well suited for evaluating evaluator reliability.

\section{Supplementary Results}
\subsection{HealthBench results}

\begin{figure}[htbp]
\centering
\includegraphics[width=\columnwidth]{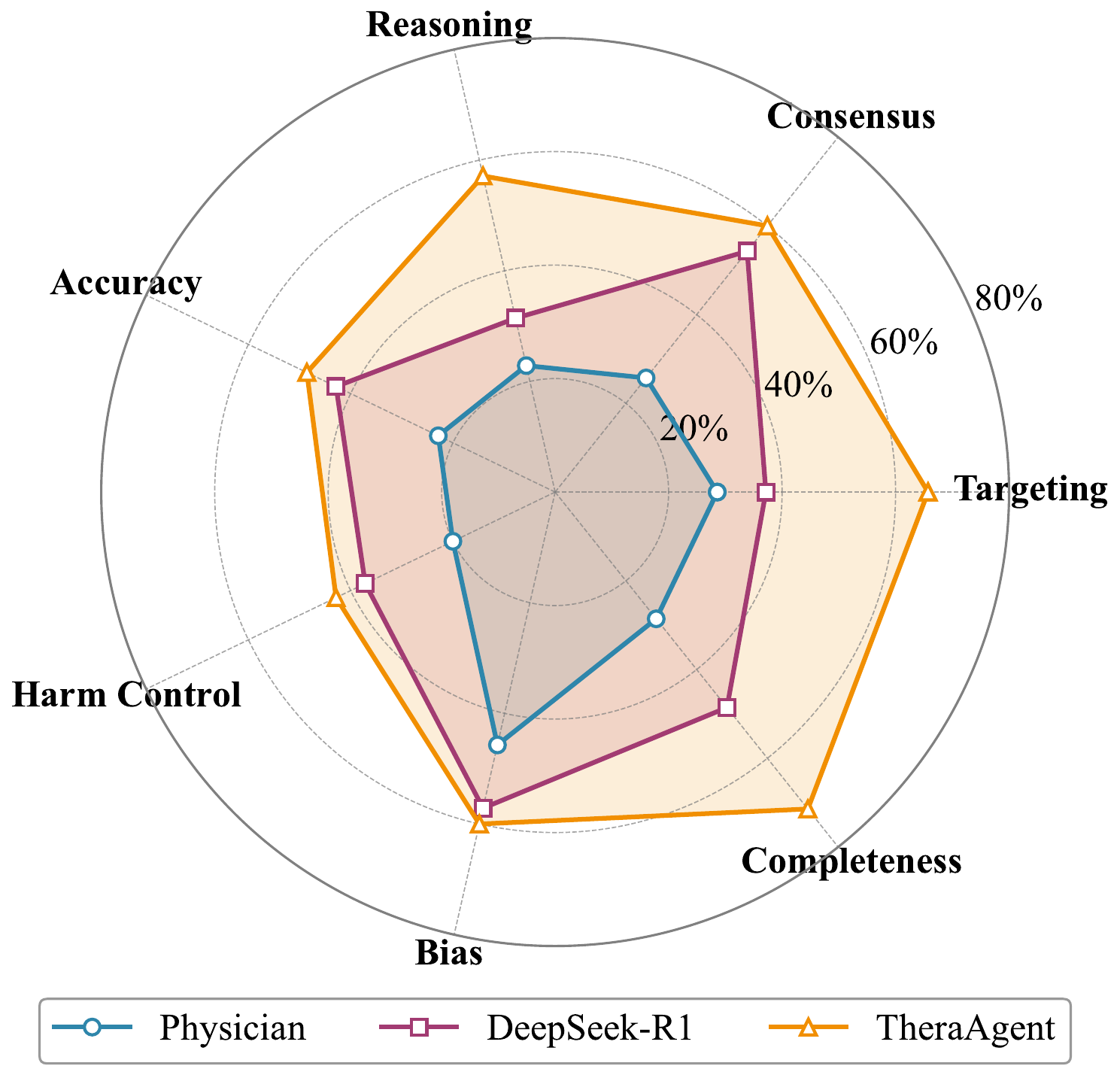}
\caption{\textbf{Comparison of high-quality rating proportions across clinical dimensions.} Data represents the percentage of expert ratings $\geq 4$ (on a 5-point scale) for all real-world medical cases.}
\label{fig:radar}
\end{figure}

\begin{figure*}[htbp]
\centering
\includegraphics[
width=\textwidth
]{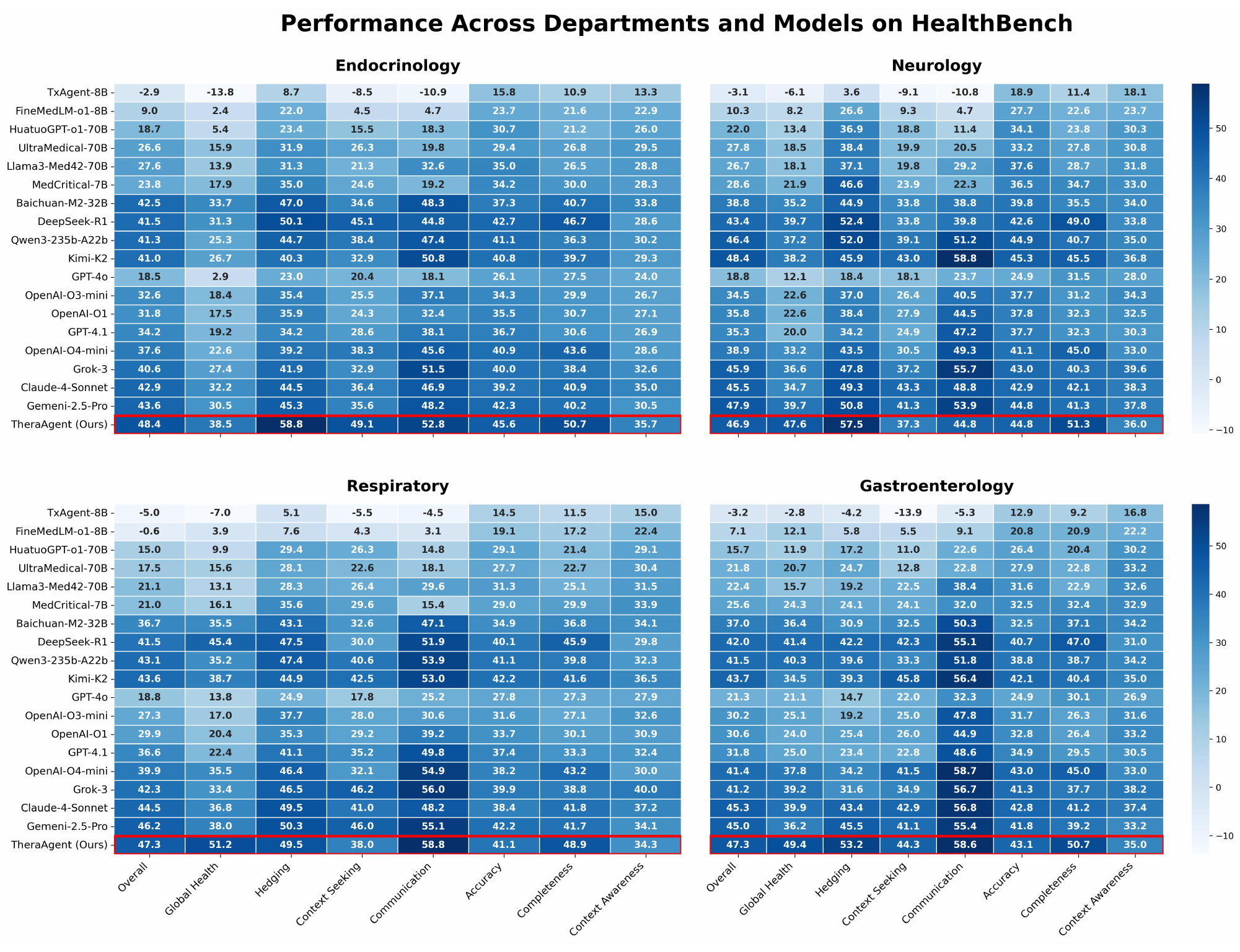}
\caption{All results on HealthBench across four departments.}
\label{fig:department_results_on_healthbench}
\end{figure*}

\label{app:all results of healthbench}
Figure~\ref{fig:department_results_on_healthbench} shows the detailed scores of TheraAgent and all baseline models on HealthBench. We report the scores on four themes: \textbf{Global Health, Hedging, Context Seeking, Communication}, three axes: \textbf{Accuracy, Completeness and Context Awareness}, as well as the overall scores. For fairness, only one experimental run is presented here.
From the results, TheraAgent consistently performs well regardless of departments. It achieves the best overall score in all but the Neurology department, where it scores 1.5 points lower than the best-performing model. 
Furthermore, TheraAgent also shows strong performance in multiple dimensions, especially on Completeness, surpassing every model in every department. These results highlight the outstanding capability of TheraAgent in ensuring completeness and avoiding critical omissions in its treatment plans, which can be attributed to its iterative refinement ability that allows continuous addition of missing information throughout the generation process.

\subsection{Rating annotations}
\label{app:rating_radar}

Figure~\ref{fig:radar} presents the proportion of high-quality ratings ($\geq 4$) across seven clinical dimensions on real-world cases. Overall, TheraAgent consistently achieves the highest scores across nearly all dimensions, indicating superior performance in generating clinically robust treatment plans. The gains are most pronounced in \textit{Targeting}, \textit{Completeness}, and \textit{Consensus}, suggesting that iterative refinement enables TheraAgent to produce more patient-specific, comprehensive, and guideline-aligned treatment plans compared to both the base model and human physicians. In contrast, DeepSeek-R1 exhibits relatively weaker performance, particularly in dimensions related to treatment precision and safety, reflecting the limitations of one-shot generation. While physician-authored regimens demonstrate strong accuracy and clinical reasoning, their lower scores in completeness and consensus adherence highlight the pragmatic abbreviations common in real-world documentation. These results further support that TheraAgent effectively balances clinical rigor, safety, and completeness, yielding more consistently high-quality therapeutic outputs.

\subsection{Ablation Study}
\label{app:ablations}

\begin{table}[t]
\centering

\resizebox{\columnwidth}{!}{%
\begin{tabular}{@{}lc@{}}
\toprule
Method & Overall Score \\ \midrule
TheraAgent w/o Memory & 0.4115 \\
- with all Memory & 0.4859 \\
- with nearest three Memory & 0.5002 \\
- with best three Memory & \textbf{0.5236} \\ \bottomrule
\end{tabular}%
}
\caption{Ablation study of Memory settings on HealthBench}
\label{memory_ablation}
\vspace{-0.5em}
\end{table}

\begin{table}[t]
\centering

\resizebox{\columnwidth}{!}{%
\begin{tabular}{@{}lc@{}}
\toprule
Method & Average Score \\ \midrule
TheraAgent w/o RAG & 89.42 \\
- inference with RAG & 90.05 \\
- inference and judge with RAG & 89.67 \\
- judge with RAG & \textbf{92.73} \\ \bottomrule
\end{tabular}%
}
\caption{Ablation study of RAG settings on the Real-World Case Dataset}
\label{RAG_ablation}
\end{table}

Table~\ref{memory_ablation} examines the effect of Dynamic Memory on HealthBench. Removing memory reduces TheraAgent to a non-iterative baseline and results in the lowest score (0.4115). Incorporating memory consistently improves performance, with selective retrieval outperforming full-memory usage. Retrieving the top three highest-scoring memory items achieves the best result (0.5236), demonstrating that score-aware and compact memory selection provides more effective guidance for iterative refinement than using all or nearest memories.

Table~\ref{RAG_ablation} investigates the effect of RAG on TheraAgent performance on Real-World Case datasets. Since both the Planner and TheraJudge can incorporate RAG, we evaluated all possible combinations where RAG is not used, used only in inference, used only in judging and used in both inference and judging. Excluding RAG completely results in the lowest score (89.42), whereas incorporating RAG in any component boosts performance, demonstrating that RAG effectively aligns treatment plans with medical consensus and enables models to generate more clinically-sound recommendations. Notably, using RAG in judging only yields the highest score (92.73), while using RAG in inference gives only marginal score increase, regardless of whether it is also used in judging.

\subsection{Case Study}
\label{app:case study}


We perform case study on a data sample arbitrarily chosen from the Real-World Case Dataset. Table~\ref{fig:case study} summarizes the given patient case information and treatment plans written by humans, DeepSeek-R1 and TheraAgent. The patient information includes symptom descriptions, medical history, test results, the final diagnosis and the diagnostic rationale. The treatment plan written by a human physician is retrieved directly from the Real-World Case dataset. The plan includes 3 sections. The Primary Treatment Section details the recommended medication and therapy. The Monitoring and Supportive Care Section lists any further assessments, long-term monitoring and medical support. The Rationale Section provides reasons for the recommended treatment. For a fair and structured comparison, DeepSeek-R1 and TheraAgent are prompted to generate a treatment plan with the same 3-section format.

\begin{table*}[t]
\begin{tcolorbox}[
  colback=blue!3,
  colframe=blue!60!black,
  title=\textbf{Case Study: CPFE Treatment Planning},
  fonttitle=\bfseries,
  boxrule=0.6pt,
  arc=1.5mm,
  toptitle=1mm
]

\small

{\normalsize\textbf{Patient Profile.}}

\textbf{Age/Sex:} 70-year-old male.

\textbf{Symptoms:} Chronic cough with sputum >4 years, worsened in last 10 days. 

\textbf{PMH:} CAD (PCI with 1 stent, 2 years ago); Type 2 diabetes (2 years); Smoking 30 years (5 cig/day, quit 2 years); No occupational/environmental exposure.

\textbf{FH:} Non-contributory. 

\textbf{PE:} T 36.5 $^\circ$C; P 78/min; RR 18/min; BP 142/70 mmHg; Alert; No JVD, cyanosis, rash, clubbing, edema; Lungs: coarse breath sounds + bibasilar crackles; Heart/abdomen normal. 

\textbf{Labs:} Hb 120 g/L; WBC $5.67×10^9/L$; Neu 0.633; Eos $0.17×10^9/L$; ESR 24 mm/h; CRP 1.6 mg/L; PCT 0.062 $\mu g/L$; D-dimer 2.54 $\mu g/L$; CEA 7.593 $\mu g/L$; SCC 4.2 $\mu g/L$; G/GM negative; Sputum bacterial/fungal/TB tests normal. 

\textbf{PFT (Feb 26, 2018):} Mild restrictive defect; Moderate ↓DLCO; ↓lung volumes and RV; Normal RV/TLC. \textbf{LCQ:} Total 17.77 (Physical 4.63, Psychological 6.14, Social 7.00). 

\textbf{Diagnosis:} CPFE. 

\textbf{Basis:} PFT showing mild restriction + moderately reduced DLCO + reduced lung volumes with normal RV/TLC; LCQ indicating significant chronic cough impact.

\vspace{0.5em}
\vspace{0.5em}

{\normalsize\textbf{Human's Treatment Plan.}}

1. \textbf{Primary Treatment:} Pirfenidone 0.4 g three times daily, with gradual dose escalation for long-term therapy. N-acetylcysteine (NAC) 0.6 g three times daily as combination antifibrotic therapy. 

2. \textbf{Monitoring and Supportive Care:} (1) Regular monitoring of liver and renal function; periodic assessment of pulmonary function and disease progression; close observation for drug-related adverse effects. (2)Provide oxygen therapy and respiratory rehabilitation guidance; ensure adequate nutritional support to maintain physical strength; deliver ongoing health education and medication adherence counseling. 

3. \textbf{Rationale:} Pirfenidone is an antifibrotic agent that slows the progression of idiopathic pulmonary fibrosis. NAC has antioxidant and antifibrotic properties; combination treatment may enhance therapeutic efficacy. Combined antifibrotic therapy helps improve prognosis and delay disease progression.

\vspace{0.5em}
\vspace{0.5em}

{\normalsize\textbf{DeepSeek-R1's Treatment Plan.}}

1. \textbf{Primary Treatment Measures:} (1)Antifibrotic therapy; Pirfenidone: Start at 200 mg orally three times daily, titrate to a target of 600 mg three times daily. Mucolytic agent: \textcolor{bad}{N-acetylcysteine 600 mg orally once daily.} (2) Management of acute exacerbations: Empiric antibiotic therapy: Amoxicillin–clavulanate 625 mg orally three times daily for 5–7 days. (3) Bronchodilators: LAMA: Tiotropium 18 $\mu g$ inhaled once daily. (4) Antitussive therapy: First-line: Dextromethorphan 10–20 mg orally 3–4 times daily. \textcolor{bad}{Second-line: Low-dose morphine, starting at 2.5 mg orally twice daily.} 

2. \textbf{Monitoring and Supportive Care:} (1) Evaluate for heart failure, rule out pulmonary embolism, and assess oxygen therapy needs. (2) Antifibrotic therapy: Monitor liver function monthly. Repeat pulmonary function tests (FVC, DLCO) after 3 months. (3) Emergency signs: Worsening cough, new-onset dyspnea, or fever—seek immediate medical care. 

3. \textbf{Rationale for Treatment:} (1) Antifibrotic agents slow lung function decline. (2) Empiric antibiotics target common lower-respiratory pathogens. (3) Bronchodilators improve symptoms related to emphysema and enhance ventilation. (4) Risks include gastrointestinal side effects of antifibrotics and morphine-related respiratory depression.

\vspace{0.5em}
\vspace{0.5em}

{\normalsize\textbf{TheraAgent's Treatment Plan.}}

1. \textbf{Major Therapeutic Measures:} (1) \textcolor{good}{Antifibrotic Therapy Decision-Making: When HRCT demonstrates a definite UIP pattern, conduct a multidisciplinary team (MDT) discussion to assess risks and benefits.} Pirfenidone: Start at 200 mg orally three times daily, titrate over 2 weeks to 600 mg three times daily. If HRCT shows a non-UIP pattern or indeterminate pattern, do not initiate antifibrotic therapy.  Antioxidant and Mucolytic Therapy: \textcolor{good}{N-acetylcysteine 600 mg orally three times daily (total 1800 mg/day).} (2) Management of Acute Exacerbation: Urgent Imaging: High-resolution chest CT (HRCT) within 24 hours. Infection Management: No empirical antibiotics. Monitor temperature and sputum daily. If purulent sputum or fever >38 °C occurs, start targeted narrow-spectrum antibiotics. (3) Bronchodilators: When HRCT confirms emphysema: First-line: Tiotropium 18 $\mu g$ inhaled once daily.  (4) Cough Suppression: First-line: .... Second-line: .... \textcolor{good}{Third-line: Morphine 1 mg orally twice daily, only when no risk of respiratory depression.}

2. \textbf{Monitoring and Supportive Care:} (1) Assess for heart failure, exclude pulmonary embolism, evaluate oxygen needs, perform arterial blood gas analysis, and screen for lung cancer. (2) Monitor liver function (ALT/AST) monthly during antifibrotic therapy. Monitor renal function (eGFR) regularly during N-acetylcysteine treatment. (3) End-Stage Planning: ... 

3. \textbf{Rationale / Evidence:} (1) HRCT is central to CPFE diagnosis and therapeutic decision-making. (2) High risk of heart failure and pulmonary embolism warrants early identification. (3) Avoid antibiotics without evidence of infection. (4) Antifibrotic therapy is restricted to HRCT-confirmed UIP patterns. (5) Bronchodilators are used only when HRCT confirms emphysema. (6) High-dose N-acetylcysteine is recommended in IPF guidelines. (7) Morphine is reserved for third-line use with strict monitoring. (8) The protocol adheres to ATS/ERS guidelines with dynamic adjustment.

\end{tcolorbox}
\caption{Detailed comparison between different methods.}
\label{fig:case study}
\end{table*}

\section{Prompts}
\subsection{Planner} 
We present the prompt template for the Planner component in Table~\ref{fig:prompt_planner}. 

\begin{table*}[t]
\begin{tcolorbox}[
  colback=gray!3,
  colframe=gray!60!black,
  title=\textbf{Prompt Template for Planner},
  fonttitle=\bfseries,
  boxrule=0.6pt,
  arc=1.5mm,
  toptitle=1mm
]

\begin{lstlisting}[style=prompt]
## `\textbf{ Patient Case Details:}`
{query}

### `\textbf{ Old treatment plan 1:}`
{treatment_plan}

### `\textbf{ Reflection to the old treatment plan 1:}`
{experience}

### `\textbf{ Score of the old treatment plan 1:}`
{score}

### `\textbf{ Old treatment plan 2:}`
{treatment_plan}

### `\textbf{ Reflection to the old treatment plan 2:}`
{experience}

### `\textbf{ Score of the old treatment plan 2:}`
{score}

...

## `\textbf{ Task:}`
You are an expert in {department}. Please think step by step to give a treatment plan for the patient accurately based on the above information.

## `\textbf{ Output format:}`
<thinking>The reasoning process</thinking>
<answer>The treatment plan</answer>
\end{lstlisting}

\end{tcolorbox}
\caption{The prompt template for the Planner component.}
\label{fig:prompt_planner}
\end{table*}

\subsection{TheraJudge} 
We present the prompt template for the TheraJudge component in Table~\ref{fig:prompt_therajudge}. Three boxed paragraphs, the \textbf{RAG context},  the \textbf{few-shot guidance} and the \textbf{multi-dimensional judging} are optional and only included when the respective functions are enabled.

\begin{table*}[t]
\begin{tcolorbox}[
  colback=gray!3,
  colframe=gray!60!black,
  title=\textbf{Prompt Template for TheraJudge},
  fonttitle=\bfseries,
  boxrule=0.6pt,
  arc=1.5mm,
  toptitle=1mm
]

\small
\begin{tcolorbox}[
  colback=gray!3,
  colframe=gray!60!black,
  title=\textbf{RAG context (if RAG is enabled)},
  fonttitle=\bfseries,
  boxrule=0.6pt,
  arc=1.5mm,
  toptitle=1mm
]
\begin{lstlisting}[style=prompt]
### `\textbf{ RAG Context:}`
{guideline 1}

{guideline 2}

...
\end{lstlisting}
\end{tcolorbox}

\begin{tcolorbox}[
  colback=gray!3,
  colframe=gray!60!black,
  title=\textbf{Few-shot guidance (if Few-shot is enabled)},
  fonttitle=\bfseries,
  boxrule=0.6pt,
  arc=1.5mm,
  toptitle=1mm
]
\begin{lstlisting}[style=prompt]
## `\textbf{Example 1:}`
### `\textbf{Example 1 Case Details:}`
{query}

### `\textbf{Example 1 Treatment Plan:}`
{treatment_plan}

### `\textbf{Example 1 Score:}`
{score}

...
\end{lstlisting}
\end{tcolorbox}

\begin{tcolorbox}[
  colback=gray!3,
  colframe=gray!60!black,
  title=\textbf{Multi-dimensional judging (if Dimension is enabled)},
  fonttitle=\bfseries,
  boxrule=0.6pt,
  arc=1.5mm,
  toptitle=1mm
]
\begin{lstlisting}[style=prompt]
`\textbf{Please evaluate the treatment plan from the following seven dimensions and give a score from 0 to 100:}`
1. Scientific Consensus Compliance (To what extent is the treatment plan consistent with established scientific and clinical consensus?)
2. Plan Completeness (To what extent does the plan comprehensively address all necessary components without omission?)
3. Situation Targeting (To what extent does the plan accurately reflect and address the patient's specific condition?)
4. Rationale-Measure Coherence (To what extent is the reasoning behind the treatment plan logically connected to the proposed measures?)
5. Harm Potential (What is the extent and likelihood of potential harm to the patient?)
6. Information Accuracy & Relevance (To what extent does the plan contain inaccurate or irrelevant information?)
7. Bias in Medical Content (To what extent does the plan exhibit bias or inapplicability to specific patient demographics?)
\end{lstlisting}
\end{tcolorbox}

\begin{lstlisting}[style=prompt]
### `\textbf{Patient Case Details:}`
{query}

### `\textbf{Treatment Plan to Evaluate: }`
{treatment_plan}



`\textbf{Please answer using the following format:}`
<reason>[detailed explanation]</reason>
<dimension_scores>[all dimension scores from 0 to 100]</dimension_scores>
<overall_score>[overall score number from 0 to 100]</overall_score>

\end{lstlisting}

\end{tcolorbox}
\caption{The prompt template for the TheraJudge component.}
\label{fig:prompt_therajudge}
\end{table*}

\end{document}